# Machine Learning for Dental Image Analysis

Young-jun Yu

Department of Computer Science

Yonsei University

aaorkr@gmail.com

## 1. Introduction

The field of pathology diagnosis has steadily advanced with the development of microscopy, accompanied by the automation of the reduction of inter-observer reliability and intra-observer reproducibility. Within the field of mammography, computer vision, and artificial intelligence (AI) techniques have been successfully applied to detect and characterize abnormalities of medical images [Winsberg et al., 1967; Ravdin et al., 2001]. This has resulted in a situation such that automated detection techniques can now implement an entire medical procedure with a high degree of accuracy. In addition, advances in computer hardware and software have increased the performance and reliability of parallel computing. The advances in this technology have, in turn, provided hardware and software advancements that are sufficiently robust to support the large computational requirements of complex Artificial Intelligence (AI) algorithms and their application to machine learning. Applications of a variety of deep learning architectures such as deep neural networks, convolutional deep neural networks, deep belief networks, and recurrent neural networks to the creation of algorithms in important fields such as natural language processing, computer vision, speech recognition and bioinformatics, have resulted in efficient and accurate automation of many pragmatic tasks [Collobert and Weston, 2008; Hinton et al., 2012; Alipanahi et al., 2015]. Using advanced large-scale parallel processing hardware and AI software, IBM developed the Watson machine to support cognitive applications across a variety of knowledge domains. Since its availability, the IBM Watson Group has developed highly reliable knowledge domains within several specialty medical fields, integrating these knowledge domains within the Watson architecture. Using these modified Watson machines, a variety of studies are in progress at the Memorial Sloan Kettering Cancer Center, the MD Anderson Cancer Center, the Cleveland Clinic, the Mayo Clinic, the New York Genome Center, the Bumrungrad International Hospital, and the Manipal Hospitals [Aggarwal and Madhukar, 2016; Herath et al., 2016; Piros et al., 2016; Raza et al., 2016]. In August 2015, IBM acquired Merge Healthcare, a company that develops a cloud-based picture archival communication system (PACS), providing IBM with the requisite authority to access the medical images of over 600 hospitals [Leidos and Pentagon, 2015]. Within six months, IBM acquired Truven Health Analytics, and gained immediate access to their 8,500 clients and medical records of nearly 300 million patients [Chang and Choi, 2016; Nash, 2016]. In September 2016, the IBM Watson Health Group announced that the Gil Medical Center of Gachon University deployed an IBM Watson machine, trained for Oncology by Memorial Sloan Kettering. While research and deployment of AI technology has been broadly conducted across the field of medicine, this is not the case in the field of dentistry. Active research of AI technology within the field of dentistry is sparse.

The purpose of this study is to evaluate the application of AI

technology to dental imaging, specifically possible deployment of artificial neural networks (ANN), a convolutional neural network (CNN), representative image cognition algorithms that support scale-invariant feature transform functions (SIFT), and the creation of histograms of oriented gradients (HOG). In this study, we conduct experiments and the results are evaluated in the following areas: accuracy (in terms of the true positive (TP) rate or recall rate) of the quantitative measures from task execution, task execution efficiency, and the visual features of the AI training. These evaluations of the CNN features were then subsequently compared to those of the image cognition algorithms (SIFT and HOG).

## 1.1. History of neural networks

Typical of most scientific fields of study, the development and application of neural networks has grown from an embryonic state to become an integral tool in many areas of applied science. This history originated in the early 1940s, at the time of the creation of the first digital computers based on the designs created by John von Neumann, such as the ENIAC and EDVAC computers, and continues in parallel with the history of the modern digital computer. The history of neural networks can be partitioned into four time periods: (1) inception, (2) the golden age, (3) the long setback period, and (4) the renaissance.

### 1.1.1. Inception

In 1943, McCulloch and Pitts presented a computational model based on a simple neural network that supported arithmetic and logical operations. Their paper strongly influenced the development of neural networks [McCulloch and Pitts, 1943]. In 1949, psychologist Donald Hebb published a book entitled "The Organization of Behavior" that presented a learning law based on neurons and synapses. Hebb applied his learning law (now known as "Hebbian learning") to explain the results of psychological experiments [Hebb, 1949].

### 1.1.2. Golden age

The 1950s and 1960s are known as the golden age of neural networks. In 1951, Minsky developed the first neurocomputer, called the Stochastic Neural Analog Reinforcement Computer (SNARC). This machine possessed the potential to automatically control the weights of synapses, but engineering the prototype into a production level machine did not occur. [Kelemen, 2007]. In 1957, Rosenblatt developed the "perceptron," an algorithm that is pervasive within an ANN. Later, he successfully developed the first neurocomputer based on the perceptron, which he applied to the field of pattern recognition [Rosenblatt, 1957]. The perceptron was expected to advance machine learning, however, its capabilities were limited. In 1969, Minsky proved that a single-layer perceptron could recognize patterns that can be divided linearly, but that complex patterns require a multi-layer ANN. Today, the perceptron is primarily used as a teaching and learning vehicle to teach binary classification algorithms. Minsky also showed that the perceptron was not able to learn sequential operations to evaluate the logical exclusive-or (XOR) function. This latter restriction eliminated the ANN as a legitimate candidate to advance machine learning.

### 1.1.3. Long setback period

During the 1970s, research funding to support ANN technology diminished, the conferences that focused on ANN technology decreased in number, and the number of published papers that addressed ANN technology also sharply decreased. However, the ANN research during this decade provided the foundation for the renaissance of ANN technology that began in the late 1980s. In 1976, Grossberg published several papers on ANNs [Grossberg, 1976]. Later, Carpenter, using Grossberg's work, developed adaptive resonance theory (ART), which, along with self-organizing maps (SOMs), made a large contribution to the field of unsupervised learning

[Carpenter, 2011].

### 1.1.4. Renaissance

In 1982 and 1984, physicist John Hopfield published two papers that encouraged the revitalization of research into ANN technology [Hopfield, 1982]. These papers were widely read, resulting in an ANN interest from a new generation of researchers. In 1986, Rumelhart and Hinton presented the back-propagation algorithm that provided solutions to many of the known ANN problems [Williams and Hinton, 1986]. The first international conference to address the renewed ANN technology was held in 1987, and the first international journal on ANNs appeared the following year. In 1995, LeCun and Bengio introduced the convolutional neural network (CNN), whose local invariant features could be easily extracted. This eliminated some of the limitations of the existing ANNs and implementations that deployed this new ANN model achieved excellent performance in the fields of character and voice recognition [LeCun and Bengio, 1995]. However, reliable results required creation of proper hyper-parameters, which in turn required an inductive algorithm. This resulted in researchers shifting focus from ANN technology in favor of the Support Vector Machine (SVM), characterized by simpler algorithms, such as linear classifiers. However, with the growth of deep learning in the late 2000s, research once again moved to ANN technology.

## 1.2. Artificial neural network

### 1.2.1. Hebbian rule

In 1949, Hebb published a learning rule based on neurons and synapses [Hebb, 1949]. He noted the fact that when learning occurs in a biological neural network, the synaptic strengths are set in order to respond well to a sign stimulus originating from a certain input. According to the Hebbian rule, learning is defined as adjusting the strength of the synaptic links. The fundamental learning method is to increase the relative synaptic weight in order to revitalize the two neurons simultaneously. The following equation captures the essence of Hebbian learning:

$$w_{ij} = \frac{1}{p} \sum_{k=1}^{p} x_i^k x_j^k$$

where $w_{ij}$ is the weight of the connection from neuron $j$ to neuron $i$, $p$ is the number of training patterns, and $x_i^k$ is the $k$th input to neuron $i$. Figure 1 shows a comparison between a human neuron and an ANN neuron [Maltarollo et al., 2013].

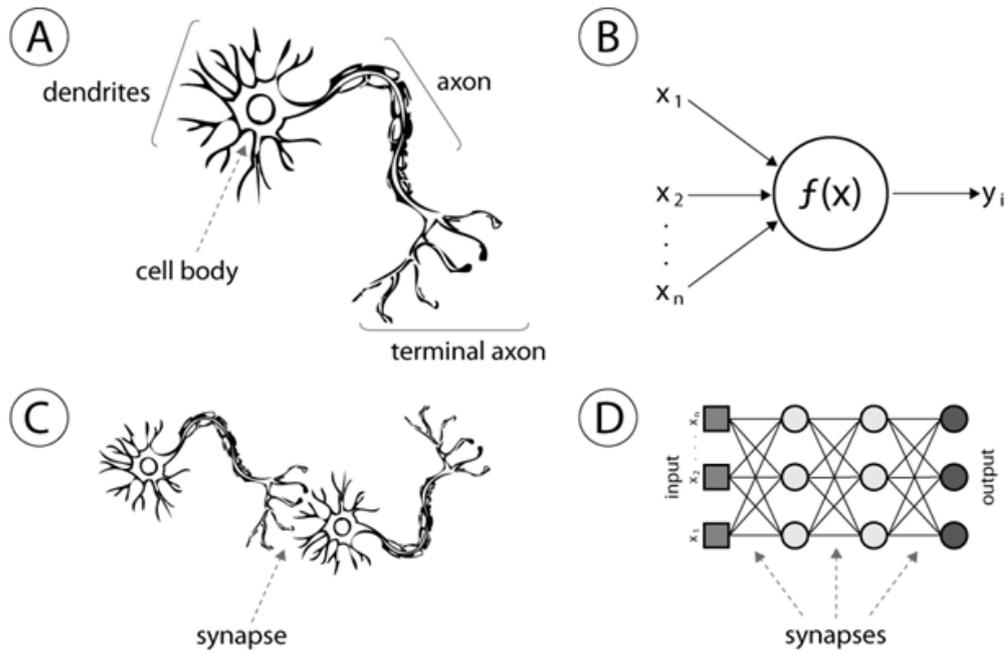

Fig. 1. (A) Human neuron; (B) Artificial neuron or hidden unity; (C) Biological synapse; (D) ANN synapses [Maltarollo et al., 2013].

### 1.2.2. Perceptron

In 1957, Rosenblatt defined the concept of a perceptron [Rosenblatt, 1957]. At the time, a neuron activation function was represented by a step function. A step function is discrete, discontinuous with a small range, and not as accurate as a continuous sigmoid function that is used today. Thus, in 1957, there were limitations in describing the behavior of a neuron. However, it was recognized that the activation of a neuron could accurately be described by a function with multiple dependent variables and a single output variable. Further, it is important that the output must be determined by the importance of the input. In this context, the importance of the input is determined by an assigned weight. Figure 2 shows an N-input perceptron with a fixed threshold. It is defined to accept N input variables, each associated with a specified weight that contribute to the computation of a unique output. The output is 0 if the sum of the product of the weights and inputs does not exceed the threshold; otherwise, the output value is 1.

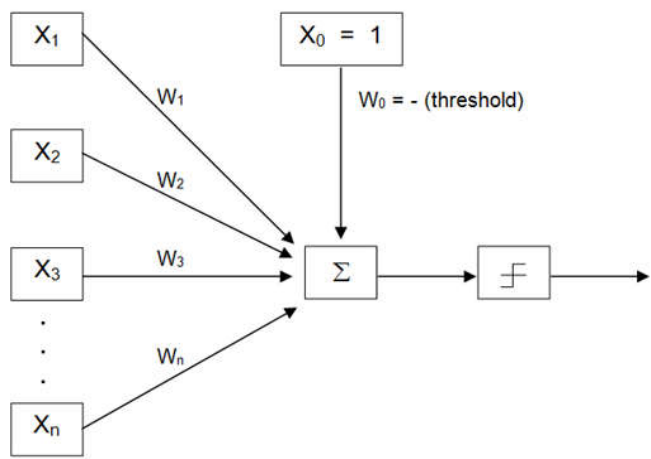

Fig. 2. An N-input perceptron with a fixed threshold that receives many inputs and determines a single output.

### 1.2.3. Perceptron limitations

While Rosenbaltt's perceptron appears to reflect the biological characteristic of a neural network, it has a serious limitation. The perceptron is capable of computing a single binary result, 0 or 1, primarily because the transfer function is a simple step function, alternating between a value of 0 and 1. If the network does not exceed two layers, then this restriction is not a problem. However, we cannot expect accurate results from a multiple-layer ANN because the granularity of the output at each layer is restricted to 0 or 1. This is simply an insufficient model for a multiple-layer ANN. To design an ANN that will generate optimum learning results, we apply back propagation and the gradient-descent method. These concepts are based on the sensitivity of a net to changes of the input and/or weight values. A small change in the input or the weight of a particular net will induce a proportionally small change in the output. The granularity of the output of a neuron based on a perceptron is large, i.e., 0 or 1, so that the behavior is discrete. This behavior is incompatible with the behavior of an ANN using back propagation in which the granularity of the output can vary and the associated learning advances in a continuous manner by making small changes in the weight and bias. This suggests that a continuous function is required to model the changes to the output as a function of the input.

### 1.2.4. Sigmoid Function

One technique to accommodate the sensitivity of output to input is to use a sigmoid function as the activation function instead of using a step function. A sigmoid function is an analytic function. This means that it is infinitely differentiable and is therefore continuous, so that small changes to the input produce small changes to the output. It is certainly capable of modeling the ANN behavior described in the previous section. As shown in Figure 3, if a sigmoid function is used as the activation function, the changes to the output values from 0 to 1 are continuous. Hence, the output can be changed in small increments when the weight or bias is changed by a small amount. It is also necessary that the range of the sigmoid function consists of the closed interval [0,1]. The sigmoid function can be formulated as follows:

$$S(t) = \frac{1}{1 + e^{-t}}$$

Here, the value of t is computed as the inner product of the input vector t $(x1, x2, x3, ...)$ and weight vector $(w1, w2, w3, ...)$ and then adding the bias. If the input is determined and there is a small change in the weight or bias (partial differential), the output changes correspondingly.

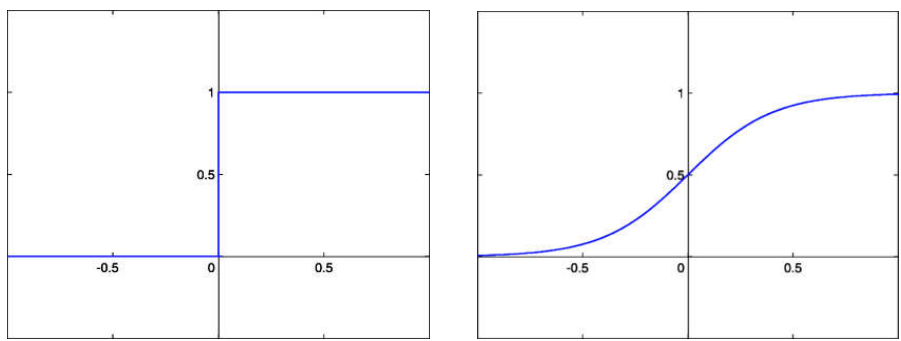

Fig. 3. Step Function (left) and sigmoid Function (right).

## 1.3. Convolutional neural network

Learning through an ANN enables us to provide solutions to many questions that are non-intuitive, such as playing the game "Go" [Silver et al., 2016]. However, in order to obtain acceptable results from a recognition algorithm based on images, applying a multi-layered neural network is difficult because this technique requires many prerequisite processes. One of the characteristics of an ANN is its requirement for a separate procedure to process new learning in the event of any change to the data. For example, a small change of a single pixel will change the size or introduce a small distortion in the image. This change is a result of the failure of the ANN to consider the topological characteristics; it processes only the raw data. Convolutional neural networks are biologically inspired variants of a multilayer perceptron designed to emulate the behavior of a visual cortex. Regarding the concept of receptive fields, the topological invariance in the local connectivity and the shared weight features will sharply reduce the number of required parameters. Convolutional neural networks were first introduced by LeCun [LeCun et al., 1989]. They were developed as part of a project to study the cognition of cursive-script zip codes. Relative to cognition of cursive-script letters, the CNNs first introduced by LeCun produced good results, but were not easily understood and thus, turned out to be ineffective as candidates for commercial production. Later, they were replaced by Benhnke [Behnke, 2003] and Simard's versions of CNN, which presented a simplification by providing an extension of the concept [Simard et al., 2003]. Later, a technique to produce CNNs was introduced through general-purpose computing algorithms for execution on graphics processing units (GPGPU).

## 2. Materials and Methods

### 2.1. Datasets and computer

We conducted experiments on panoramic dental radiographs to evaluate the application of the AI technology in dental imaging. A panoramic radiograph is a panoramic scanning dental X-ray of the upper and lower jaws. It shows a two-dimensional view of a half-circle from ear to ear. The subjects of this study were randomly selected from 972 patients aged 20 years and older who visited the Pusan National University Hospital between 2014 and 2015. The patients comprised 543 men and 429 women with a mean age of 25.3 years (range: 20-40 years). All panoramic radiographs were obtained with a Proline XC machine (Planmeca Co., Helsinki, Finland). This study was reviewed and approved by the institutional review board of the Pusan National University Dental Hospital (PNUPH-2015-034). We used up to eight CPU cores (at 2.8 GHz) and a GeForce GTX 980 GPU. All the experiments described below were executed on a single machine.

### 2.2. Bag-of-Words models

In this work, using SIFT, HOG2×2, HOG3×3, and Color features as image descriptors, we applied the Bag-of-Words (BoW) and a spatial pyramid pipeline. Subsequently, two different classifiers were deployed, specifically the k-nearest neighbors algorithm (k-NN) and an error-correcting output coding support vector machine (ECOC-SVM). Figure 4 shows the overview of our system for Bag-of-Words. All computations generated by the Bag-of-Words models were performed using the MATLAB R2016a environment. MATLAB is a numerical computing environment developed by MathWorks. It supports a fourth-generation programming language, allowing user extensions implemented in C, C++, and Python. It has the capability to create sophisticated Mathematical algorithms. Its "parfor" command was used to implement execution of parallel loop iterations [Luszczek, 2009].

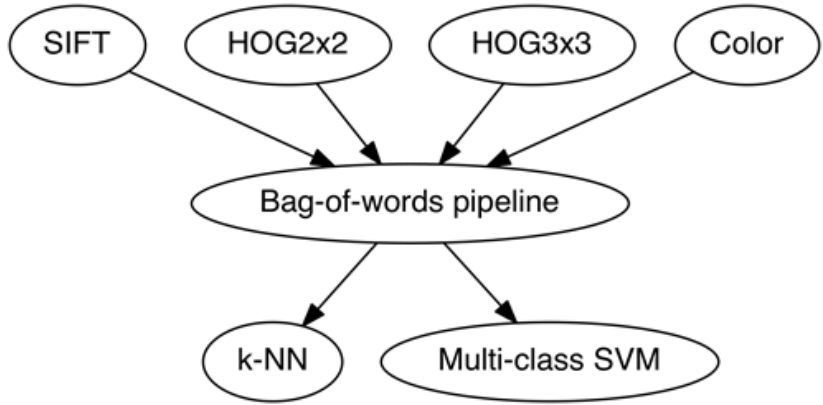

Fig. 4. Overview of our system for Bag-of-Words.

### 2.2.1. SIFT

A scale-invariant feature transform is a computer vision algorithm that can detect and describe local features in images. This study used a method for extracting distinctive invariant features from images that can be used to perform reliable matching between different views of an object or scene [Lowe, 2004].

### 2.2.2. HOG2×2 and HOG 3×3

The histogram of oriented gradients is similar to the scale-invariant feature transform descriptors; however, it differs in that it is computed on a dense grid of uniformly spaced cells and uses overlapping local contrast normalization. After reviewing existing edge and gradient-based descriptors, this study concluded that grids of Histograms of Oriented Gradient (HOG) descriptors significantly outperformed existing feature sets for detection [Dalal, 2005; Russell et al., 2008].

### 2.2.3. Color

As a method to extract essential features, the image was converted to color names, thereby enabling extraction of dense overlapping patches of multiple sizes; the patches were then aggregated to form a histogram of color names [Weijer et al., 2009; Khan et al., 2013].

### 2.2.4. Bag-of-Words pipeline

The Bag-of-Words (BoW) model can be applied to image classification by treating image features as words. A bag of visual words is a vector of occurrence counts of elements of a vocabulary of local image features. Using a random sampling of the extracted features from various patches, the k-means learning algorithm was applied to learn a dictionary [Elkan, 2003]. This was followed by the application of LLC to soft-encode each patch to one or more dictionary entries [Wang et al., 2010]. We then applied max pooling with a spatial pyramid to obtain the final feature vector [Lazebnik et al., 2006; Wang et al., 2010]. We used LLC because it supports classification by the use of a linear classifier so that it is not necessary to use the more complicated nonlinear kernels.

### 2.2.5. k-nearest neighbors algorithm

The k-NN algorithm is a non-parametric classification technique. The output is a class membership specification. An object is classified by a majority vote of its neighbors, with the object being assigned to the class most common among its k nearest neighbors.

### 2.2.6. Multi-class SVM

A Support Vector Machine (SVM) is a supervised learning model with associated learning algorithms that analyze data, recognize patterns, and, based on an identified pattern, assigns a classification. SVMs are deterministic binary linear classifiers and can perform nonlinear classification by implicitly mapping inputs into large-dimension feature spaces using kernel methods. Classification was performed with the machine learning toolbox available in MATLAB. The "fitcecoc" method, which fits multiclass models for support vector machines or other classifiers, was employed. The classifier was used with both linear and Gaussian kernels, and k-fold cross-validation was applied for classifier assessment. [Tan et al., 2016]

## 2.3. Convolutional neural network models

Figure 5 shows an overview of our system for a CNN. This study trained CNNs using Theano, which is a Python library that allows mathematical expressions to be defined, optimized, and evaluated efficiently in terms of multi-dimensional arrays. It optimizes a user's symbolically specified mathematical computations to produce efficient low-level implementations [Bergstra et al., 2010]. The size of the original panoramic radiograph was 2800 × 1376 pixels, and the CNN training is slow if the input is large. Therefore, cropped images were used to classify the teeth, and resized images were used to classify the sex as determined from panoramic radiographs. Two types of convolutional neural networks were used, 4-layer and 16-layer CNN models. As shown in Figure 6, the 4-layer CNN model is a simple stack consisting of two convolution layers and a max-pooling layer, providing an architecture similar to the architectures that Yann LeCun advocated in the 1990s for images, speech, and time series [LeCun and Bengio, 1995]. In our study, the only difference is the choice of activation function. In this study, we elected to use rectified linear unit (ReLU) activation functions rather than sigmoid activation functions. There are two major advantages of ReLU activation functions: (1) their sparsity and (2) reduced vanishing gradient. The constant gradient of the ReLU activation function results in faster learning [Nair and Hinton, 2010]. The final layer deployed either a sigmoid or softmax activation function. As shown in Figure 7, the 16-layer CNN model is an improved version of the model used by the VGG team in the ILSVRC-2014 competition [Simonyan and Zisserman, 2014]. It was obtained by directly converting the Caffe model provided by the authors. The loss functions were binary_crossentropy and categorical_crossentropy, the preferred logarithmic loss functions for classification problems. The equation that defines the description of crossentropy is as follows:

$$crossentropy(t,o) = -(t \cdot \log(o) + (1-t) \cdot \log(1-o))$$

The gradient descent optimization functions utilized were adadelta and rmsprop. The weight initialization was uniformly distributed. The scale was a uniform distribution of each datum between -0.05 and 0.05. A k-fold cross-validation was used for classifier assessment [Tan et al., 2016].

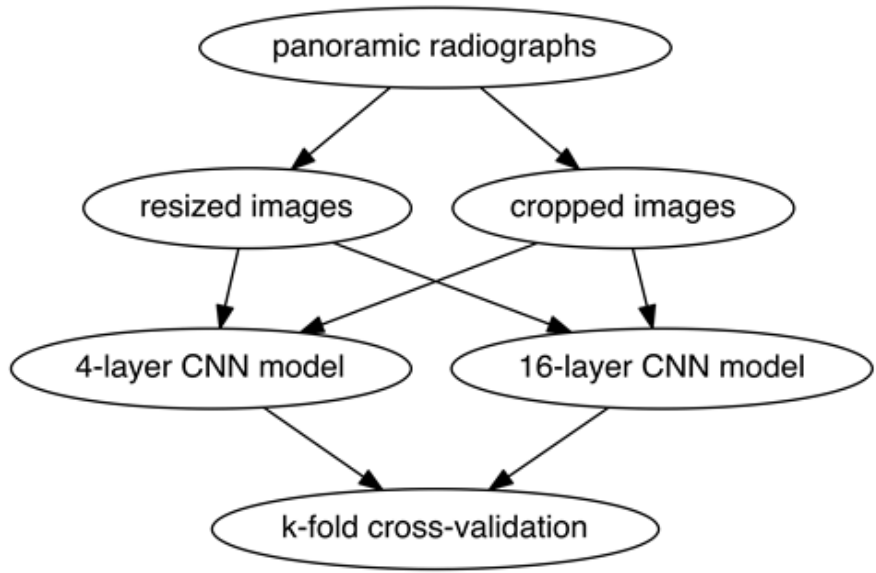

Fig. 5. Overview of our CNN system.

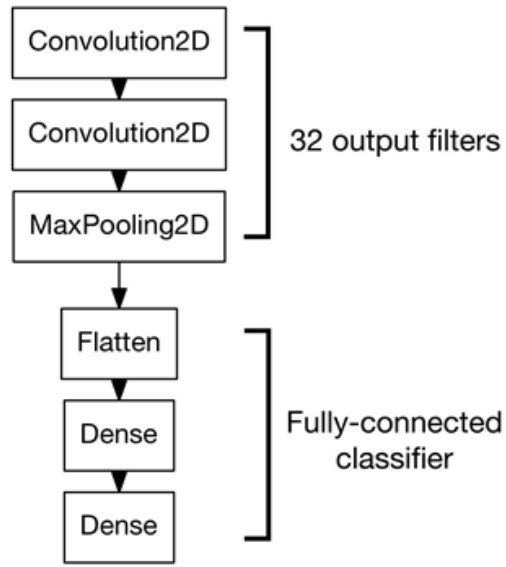

Fig. 6. 4-layer CNN model.

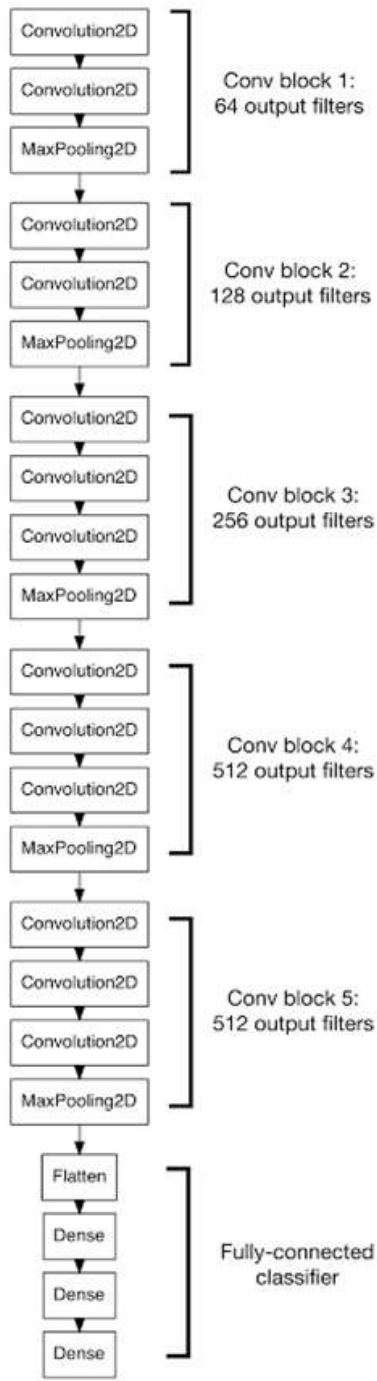

Fig. 7. 16-layer CNN model.

# 3. Results

## 3.1. Teeth Classification

We presented the AI learning tool with 100 images of 28 permanent teeth (2,800 images in total), we then classified 1,120 images of the 28 permanent teeth that were not used in the learning. To build the 28-way classifier, we adopted the FDI (FDI World Dental Federation) two-digit system that was introduced in 1970 by the FDI. It is a digital system that notates teeth that make visual sense, cognitive sense, and computer sense [Peck and Peck, 1996]. The FDI method identifies each of the 32 permanent teeth with a two-digit number, the first digit indicating the quadrant (1 to 4) while the second digit designates the tooth type (1 to 8). In this study, third molars were excluded from tooth classification in order to obtain more training samples because there are many people without third molars. A sample of 140 patients with 28 healthy permanent teeth (except four third molars from the 32 permanent teeth) was selected from 972 patients. We divided the dataset into a training set and a test set. The training set was comprised of 55 men and 45 women, and the test set was comprised of 20 men and 20 women.

### 3.1.1. Input Layer

As illustrated in Figure 8, we cropped the image of each tooth to a minimum rectangle that included the crown and root. In total, 3,920 cropped images were used (28 permanent teeth times 140 people). The average width of the cropped images was $131.6 \pm 43.73$ pixels and the average height was $282.16 \pm 37.11$ pixels. In order to find a proper training input size, the cropped images were first resized to a width and height of 32 × 32 pixels. Subsequently, images of sizes 64 × 64 pixels, 128 × 128 pixels, and 256 × 256 pixels were examined. As shown in Figure 9, for dictionary sizes of more than 500, the images with a size of 128 × 128 pixels showed the highest mean accuracy. Therefore, for the purposes of this study, we fixed the size of input images to 128 × 128 pixels.

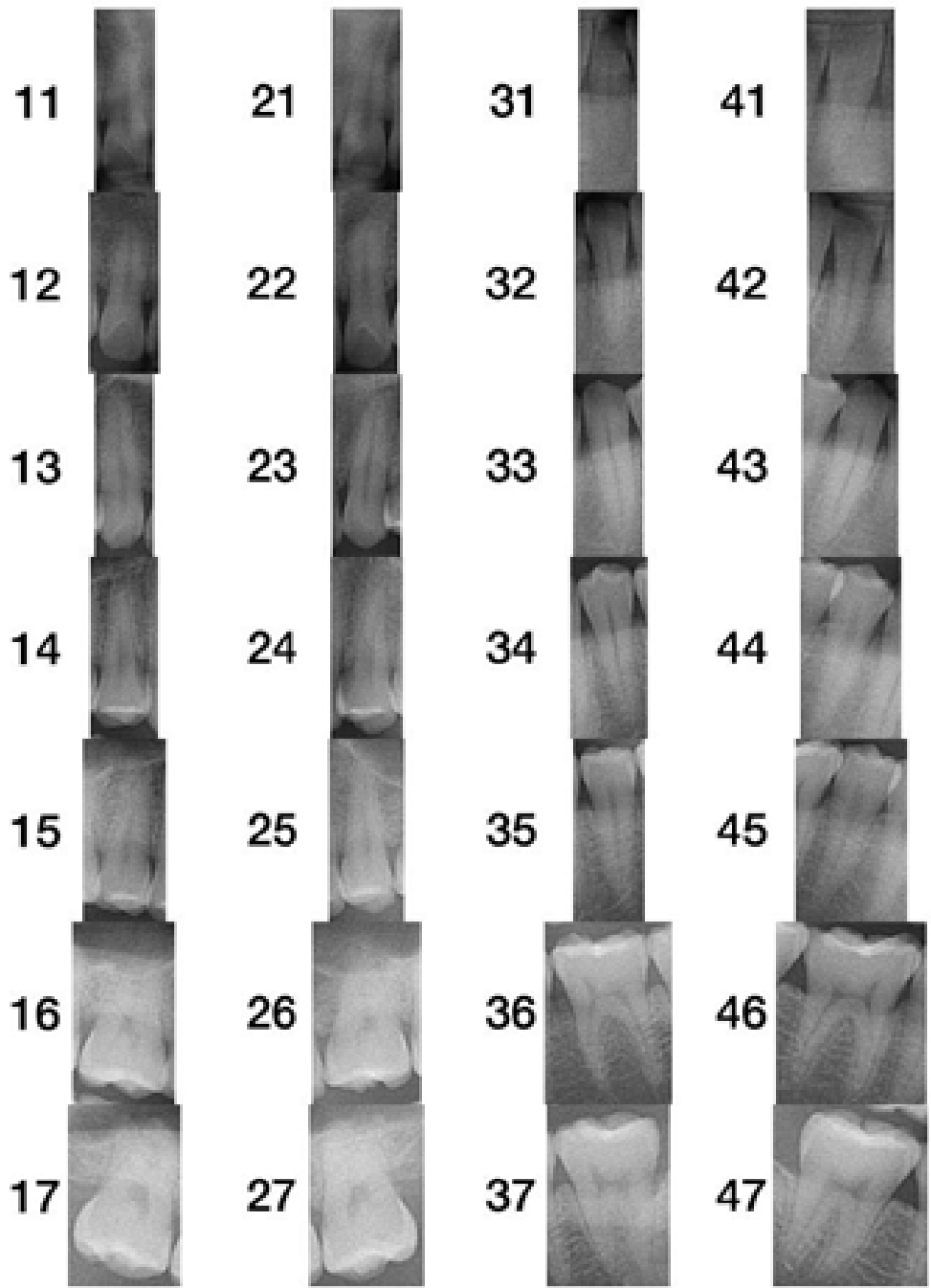

Fig. 8. Cropped images of 28 permanent teeth.

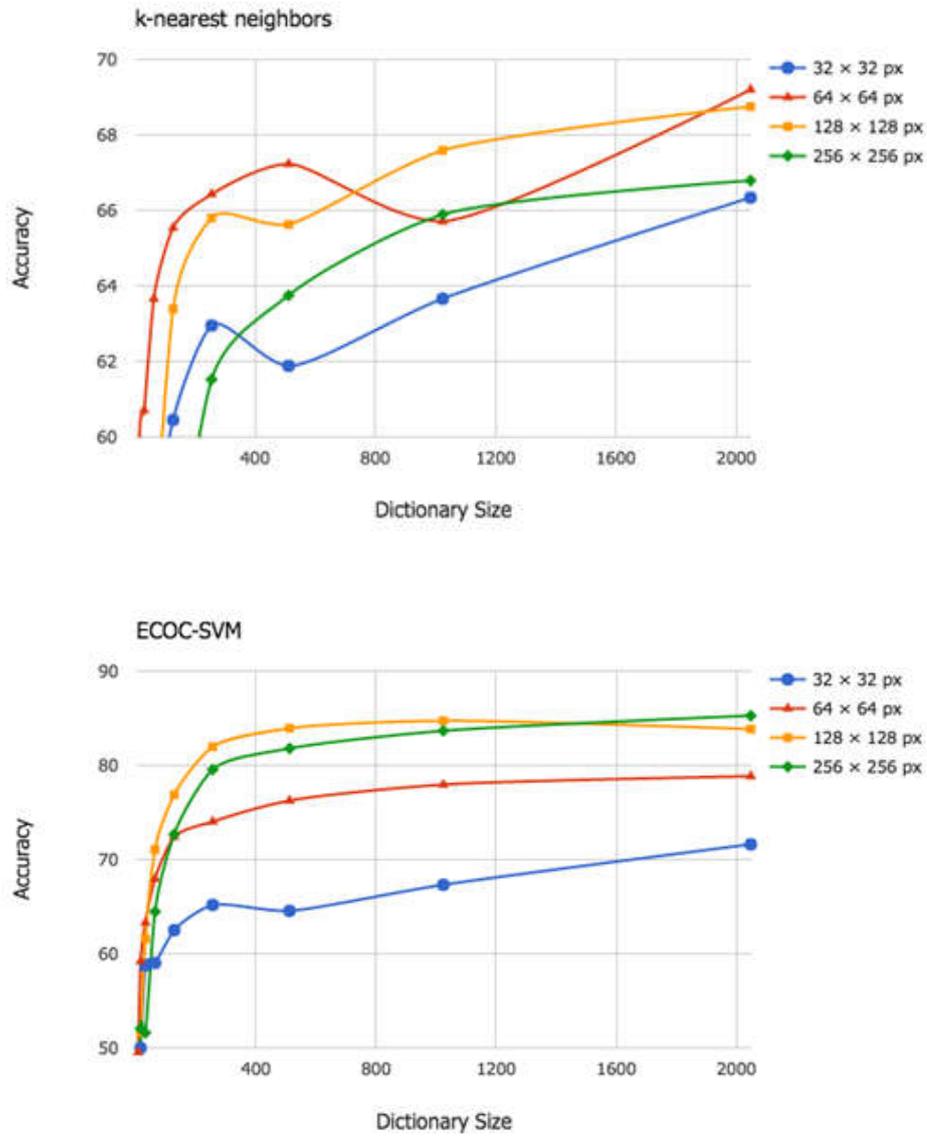

Fig. 9. Changes in accuracy with dictionary size with various input sizes using the SIFT algorithm: classified by k-NN (top) and classified by ECOC-SVM (bottom).

### 3.1.2. Spatial Pyramid Level

One of the disadvantages of the BoW model is that it ignores the spatial relationships among the patches, which are very important in image representation. Researchers have proposed several methods to incorporate the spatial information. We applied a max pooling technique with a spatial pyramid to obtain the final feature vector [Lazebnik et al., 2006]. Figure 10 shows an example of constructing a three-level pyramid. In order to find a proper spatial pyramid level, pyramid levels 0     were examined. As shown in Figure 11, for dictionary sizes of more than 500, pyramid level 2 showed the highest accuracy. Therefore, within the present study, the pyramid level was fixed to level 2.

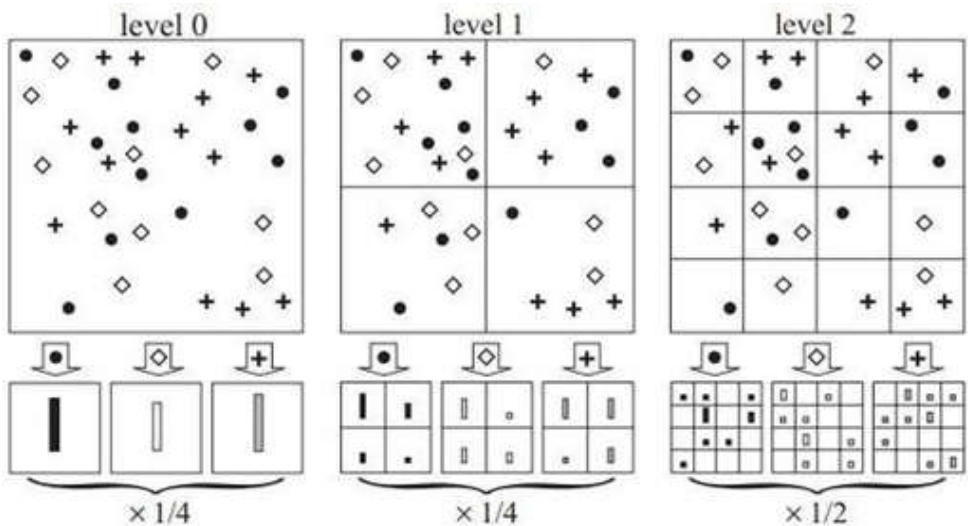

Fig. 10. Sample example of constructing a three-level pyramid. The image has three feature types, indicated by circles, diamonds, and crosses [Lazebnik et al., 2006].

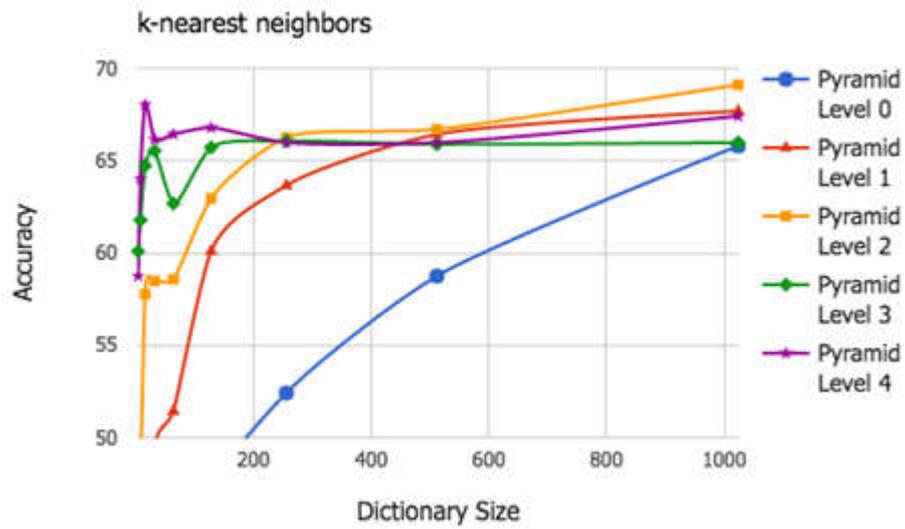

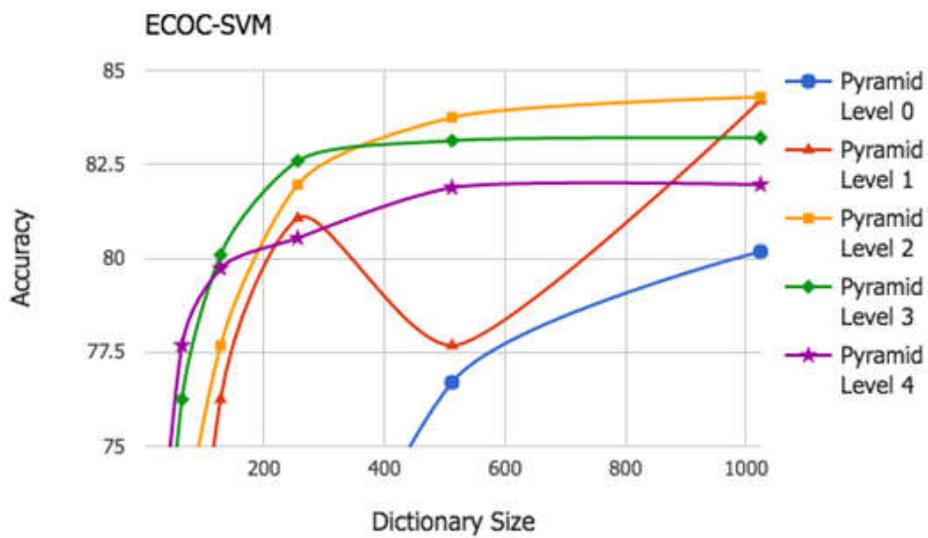

Fig. 11. Changes in accuracy with dictionary size with various pyramid levels using SIFT algorithm: classified by k-NN (top) and classified by ECOC-SVM (bottom).

### 3.1.3. Tests on Bag-of-Words Model

Figure 12 shows changes in the accuracy with a dictionary size of 128 × 128 pixels input size and a two-level pyramid using SIFT, HOG2×2, and HOG3×3 in conjunction with the Color algorithm. The result produced a maximum accuracy of 84.73% in the case of SIFT, 84.82% in the case of HOG2×2, 84.64% in the case of HOG2×2, and 65.8% in the case of Color. From the accuracy rates above, it required 283.44 s to distinguish 1,120 teeth images with SIFT, 279.95 s with HOG2×2, 110.30 s with HOG3×3, and 123.53 s with Color.

### 3.1.4. Tests on 4-layer CNN model and 16-layer CNN model

Figure 13 shows changes in accuracy using a number of training images at 128 × 128 pixels input size generated by a convolutional neural network. In the 4-layer CNN model, the mean training time over 2,800 teeth images was 4.71 s per number of training images, whereas using the 16-layer CNN model, the mean training time was 47.52 s per number of training images. With the 4-layer CNN model, distinguishing 1,120 images required 1.11 s, whereas with the 16-layer CNN model the time was 4.27 s. Note with the 4-layer CNN model, the training time was shorter but the accuracy was larger with the 16-layer CNN model. After 40 training iterations, the accuracy rate was over 80% for the test set; after 232 training iterations, the accuracy rate was over 90%. As shown in Figure 14, the accuracy rate increased with the increase in the training size. Higher accuracy is expected with more training data.

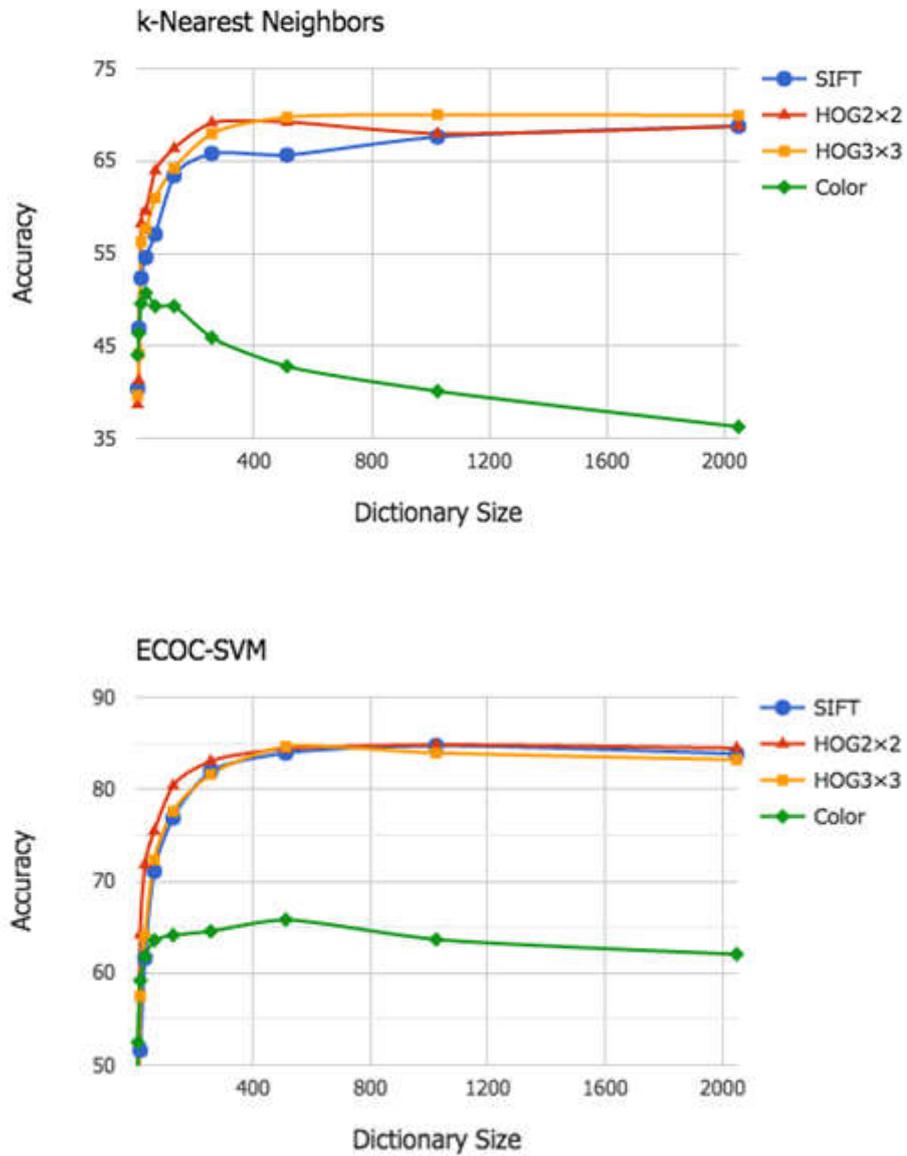

Fig. 12. Changes in accuracy with dictionary size at 128 × 128 pixels input size and level-2 pyramid using SIFT, HOG2×2, HOG3×3, and the Color algorithm: classified by k-NN algorithm (top) and classified by ECOC-SVM algorithm (bottom).

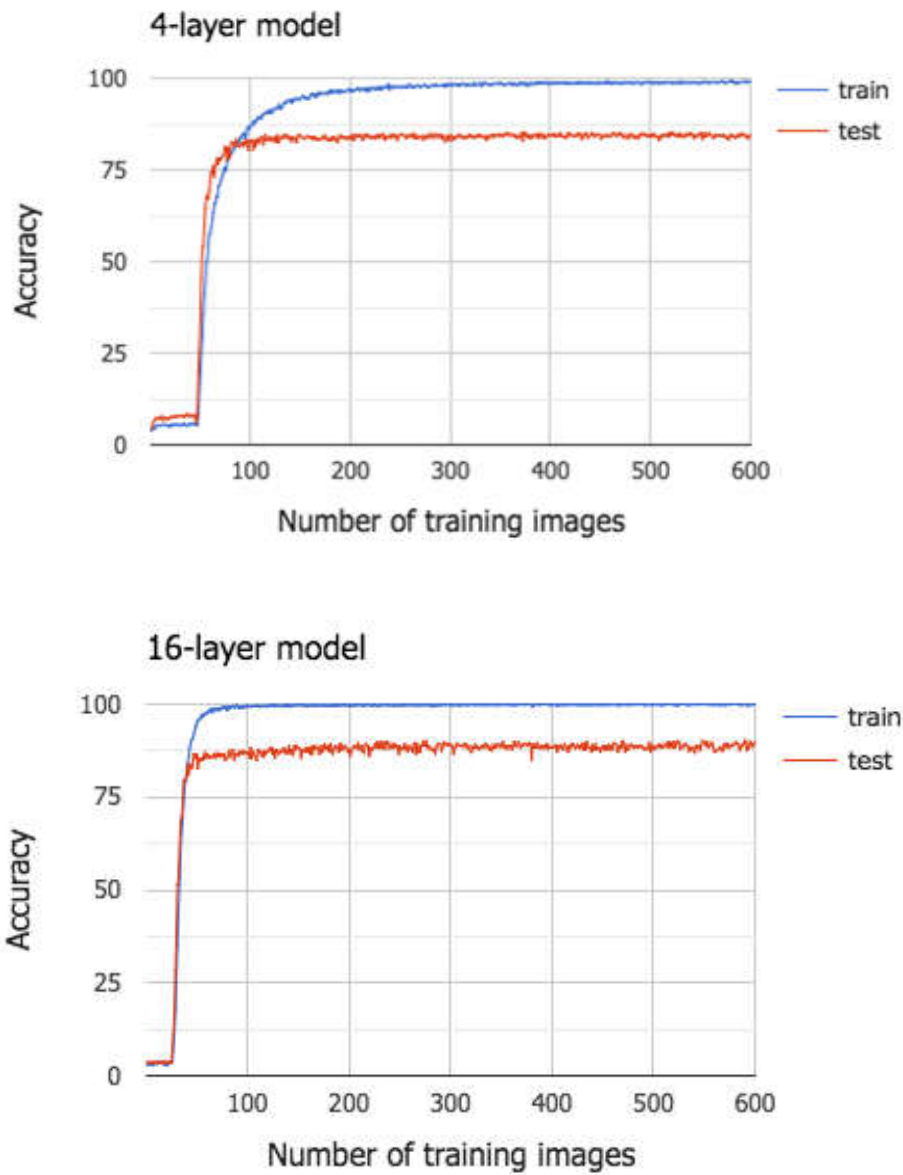

Fig. 13. Changes in accuracy with a number of training images at 128 × 128 pixels input size using the convolutional neural network: classified by a 4-layer CNN model (top) and a 16-layer CNN model (bottom).

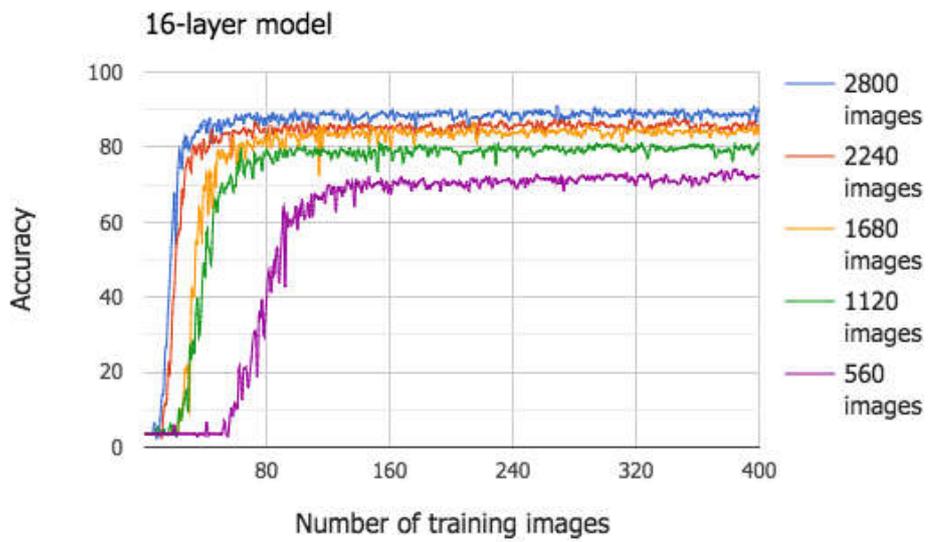

Fig. 14. Changes in accuracy with a number of training images with different training sizes using a 16-layer convolutional neural network model.

### 3.1.5. k-fold Cross-Validation on 16-layer CNN Model

There were not enough data available to partition a 16-layer CNN model into separate training and test sets without losing testing capability. In these cases, a fair way to properly estimate model prediction performance is to use cross-validation. A k-fold cross-validation of the 16-layer CNN model for teeth classification was performed with k = 14 and a total of 3,920 images (trained on 3,640 images, validated on 280 images). As shown in Table 1, molar teeth including teeth #16 and #17 obtained a high F1 score. Teeth #13 and #23 (maxillary incisor) also achieved a high F1 score. Teeth #41 and #31 (mandibular central incisor) produced the lowest F1 score. In general, maxillary teeth achieved a higher F1 score than the F1 scores of mandibular teeth. Table 2 shows a confusion matrix of the 28-way classifier over the validation dataset. Each row $i$ represents the number count for category type $i$, element $ij$ is the number of images incorrectly classified as category $j$, and element $ii$ is the total count number of images correctly classified as the correct class type $i$. The cases such that tooth #41 was mistaken for tooth #42 and tooth #42 was mistaken for tooth #41 occurred most frequently (21 times). The second most frequent misclassification occurred when tooth #14 was misclassified as tooth #15 (19 times). The third most frequent case occurred when (a) tooth #24 was misclassified as tooth #25, (b) tooth #31 was misclassified as tooth #41, and (c) tooth #41 was misclassified as tooth #31 (16 times).

Table 1. Precision and recall results for teeth classification. Condition positive (CP) of each tooth is 140 (k-fold cross-validation of 16-layer CNN model with k=14)

| Tooth number | Predicted condition positive (PCP) | True positive (TP) | Precision (TP/PCP) | Recall (TP/CP) | F1 score |
|---|---|---|---|---|---|
| 16 | 140 | 136 | 0.9714 | 0.9714 | 0.9714 |
| 17 | 140 | 136 | 0.9714 | 0.9714 | 0.9714 |
| 13 | 139 | 135 | 0.9712 | 0.9643 | 0.9677 |
| 23 | 136 | 133 | 0.9779 | 0.95 | 0.9637 |
| 36 | 138 | 133 | 0.9638 | 0.95 | 0.9569 |
| 46 | 139 | 133 | 0.9568 | 0.95 | 0.9534 |
| 37 | 140 | 133 | 0.95 | 0.95 | 0.95 |
| 47 | 138 | 131 | 0.9493 | 0.9357 | 0.9425 |
| 27 | 143 | 133 | 0.9301 | 0.95 | 0.9399 |
| 26 | 139 | 131 | 0.9424 | 0.9357 | 0.939 |
| 11 | 140 | 130 | 0.9286 | 0.9286 | 0.9286 |
| 43 | 143 | 131 | 0.9161 | 0.9357 | 0.9258 |
| 35 | 150 | 134 | 0.8933 | 0.9571 | 0.9241 |
| 44 | 143 | 130 | 0.9091 | 0.9286 | 0.9187 |
| 21 | 137 | 127 | 0.927 | 0.9071 | 0.9169 |
| 45 | 138 | 127 | 0.9203 | 0.9071 | 0.9137 |
| 12 | 138 | 125 | 0.9058 | 0.8929 | 0.8993 |
| 34 | 137 | 124 | 0.9051 | 0.8857 | 0.8953 |
| 22 | 141 | 125 | 0.8865 | 0.8929 | 0.8897 |
| 24 | 137 | 122 | 0.8905 | 0.8714 | 0.8808 |
| 33 | 134 | 119 | 0.8881 | 0.85 | 0.8686 |
| 14 | 136 | 119 | 0.875 | 0.85 | 0.8623 |
| 25 | 147 | 123 | 0.8367 | 0.8786 | 0.8571 |
| 15 | 147 | 122 | 0.8299 | 0.8714 | 0.8501 |
| 32 | 141 | 103 | 0.7305 | 0.7357 | 0.7331 |
| 42 | 143 | 102 | 0.7133 | 0.7286 | 0.7209 |
| 31 | 140 | 100 | 0.7143 | 0.7143 | 0.7143 |
| 41 | 136 | 89 | 0.6544 | 0.6357 | 0.6449 |

Table 2. Confusion matrix for the 28 way classifier of validation dataset. Each row *i* represents the number count for the category type *i*, element *ij* is the number of images misclassified as category *j*, and element *ii* is the total count number of images correctly classified as true class type *i* (k-fold cross-validation of 16-layer CNN model for teeth classification with k=14; 3,920 images)

| | 11 | 12 | 13 | 14 | 15 | 16 | 17 | 21 | 22 | 23 | 24 | 25 | 26 | 27 | 31 | 32 | 33 | 34 | 35 | 36 | 37 | 41 | 42 | 43 | 44 | 45 | 46 | 47 |
|---|---|---|---|---|---|---|---|---|---|---|---|---|---|---|---|---|---|---|---|---|---|---|---|---|---|---|---|---|
| 11 | 130 | 5 | 0 | 0 | 0 | 0 | 0 | 2 | 2 | 0 | 0 | 0 | 0 | 1 | 0 | 0 | 0 | 0 | 0 | 0 | 0 | 0 | 0 | 0 | 0 | 0 | 0 | 0 |
| 12 | 4 | 125 | 2 | 1 | 0 | 0 | 0 | 1 | 4 | 0 | 0 | 1 | 0 | 0 | 0 | 1 | 0 | 0 | 0 | 0 | 0 | 1 | 0 | 0 | 0 | 0 | 0 | 0 |
| 13 | 1 | 1 | 135 | 0 | 1 | 0 | 0 | 0 | 1 | 0 | 1 | 0 | 0 | 0 | 0 | 0 | 0 | 0 | 0 | 0 | 0 | 0 | 0 | 0 | 0 | 0 | 0 | 0 |
| 14 | 0 | 0 | 1 | 119 | 19 | 0 | 0 | 0 | 0 | 0 | 0 | 0 | 0 | 0 | 1 | 0 | 0 | 0 | 0 | 0 | 0 | 0 | 0 | 0 | 0 | 0 | 0 | 0 |
| 15 | 0 | 0 | 1 | 14 | 122 | 2 | 0 | 0 | 0 | 1 | 0 | 0 | 0 | 0 | 0 | 0 | 0 | 0 | 0 | 0 | 0 | 0 | 0 | 0 | 0 | 0 | 0 | 0 |
| 16 | 0 | 0 | 0 | 0 | 1 | 136 | 3 | 0 | 0 | 0 | 0 | 0 | 0 | 0 | 0 | 0 | 0 | 0 | 0 | 0 | 0 | 0 | 0 | 0 | 0 | 0 | 0 | 0 |
| 17 | 0 | 0 | 0 | 0 | 1 | 2 | 136 | 0 | 0 | 0 | 1 | 0 | 0 | 0 | 0 | 0 | 0 | 0 | 0 | 0 | 0 | 0 | 0 | 0 | 0 | 0 | 0 | 0 |
| 21 | 2 | 2 | 0 | 0 | 0 | 0 | 0 | 127 | 7 | 1 | 0 | 0 | 0 | 0 | 0 | 0 | 0 | 0 | 0 | 0 | 0 | 0 | 1 | 0 | 0 | 0 | 0 | 0 |
| 22 | 2 | 5 | 0 | 1 | 0 | 0 | 0 | 6 | 125 | 1 | 0 | 0 | 0 | 0 | 0 | 0 | 0 | 0 | 0 | 0 | 0 | 0 | 0 | 0 | 0 | 0 | 0 | 0 |
| 23 | 0 | 0 | 0 | 0 | 0 | 0 | 0 | 0 | 2 | 133 | 1 | 3 | 0 | 0 | 0 | 0 | 0 | 0 | 0 | 0 | 0 | 0 | 0 | 1 | 0 | 0 | 0 | 0 |
| 24 | 0 | 0 | 0 | 1 | 0 | 0 | 0 | 0 | 0 | 0 | 122 | 16 | 0 | 0 | 0 | 0 | 0 | 0 | 0 | 0 | 0 | 0 | 1 | 0 | 0 | 0 | 0 | 0 |
| 25 | 0 | 0 | 0 | 0 | 3 | 0 | 0 | 0 | 0 | 0 | 12 | 123 | 1 | 1 | 0 | 0 | 0 | 0 | 0 | 0 | 0 | 0 | 0 | 0 | 0 | 0 | 0 | 0 |
| 26 | 0 | 0 | 0 | 0 | 0 | 0 | 0 | 0 | 0 | 0 | 0 | 2 | 131 | 7 | 0 | 0 | 0 | 0 | 0 | 0 | 0 | 0 | 0 | 0 | 0 | 0 | 0 | 0 |
| 27 | 0 | 0 | 0 | 0 | 0 | 0 | 0 | 0 | 0 | 0 | 0 | 1 | 6 | 133 | 0 | 0 | 0 | 0 | 0 | 0 | 0 | 0 | 0 | 0 | 0 | 0 | 0 | 0 |
| 31 | 0 | 0 | 0 | 0 | 0 | 0 | 0 | 1 | 0 | 0 | 0 | 0 | 0 | 0 | 100 | 13 | 0 | 0 | 0 | 0 | 0 | 16 | 7 | 3 | 0 | 0 | 0 | 0 |
| 32 | 0 | 0 | 0 | 0 | 0 | 0 | 0 | 0 | 0 | 0 | 0 | 0 | 0 | 0 | 18 | 103 | 10 | 0 | 0 | 0 | 0 | 6 | 3 | 0 | 0 | 0 | 0 | 0 |
| 33 | 0 | 0 | 0 | 0 | 0 | 0 | 0 | 0 | 0 | 0 | 0 | 0 | 0 | 0 | 0 | 10 | 119 | 4 | 1 | 0 | 1 | 2 | 3 | 0 | 0 | 0 | 0 | 0 |
| 34 | 0 | 0 | 0 | 0 | 0 | 0 | 0 | 0 | 0 | 0 | 0 | 0 | 0 | 0 | 0 | 0 | 2 | 124 | 14 | 0 | 0 | 0 | 0 | 0 | 0 | 0 | 0 | 0 |
| 35 | 0 | 0 | 0 | 0 | 0 | 0 | 0 | 0 | 0 | 0 | 0 | 0 | 0 | 0 | 0 | 0 | 0 | 6 | 134 | 0 | 0 | 0 | 0 | 0 | 0 | 0 | 0 | 0 |
| 36 | 0 | 0 | 0 | 0 | 0 | 0 | 0 | 0 | 0 | 0 | 0 | 0 | 0 | 0 | 0 | 0 | 0 | 0 | 1 | 133 | 5 | 0 | 0 | 0 | 0 | 1 | 0 | 0 |
| 37 | 0 | 0 | 0 | 0 | 0 | 1 | 0 | 0 | 0 | 0 | 1 | 0 | 0 | 0 | 0 | 0 | 0 | 0 | 0 | 5 | 133 | 0 | 0 | 0 | 0 | 0 | 0 | 0 |
| 41 | 1 | 0 | 0 | 0 | 0 | 0 | 0 | 0 | 0 | 0 | 0 | 0 | 0 | 0 | 16 | 8 | 1 | 2 | 0 | 0 | 0 | 89 | 21 | 1 | 1 | 0 | 0 | 0 |
| 42 | 0 | 0 | 0 | 0 | 0 | 0 | 0 | 0 | 0 | 0 | 0 | 0 | 0 | 0 | 6 | 4 | 2 | 0 | 0 | 0 | 0 | 21 | 102 | 5 | 0 | 0 | 0 | 0 |
| 43 | 0 | 0 | 0 | 0 | 0 | 0 | 0 | 0 | 0 | 0 | 0 | 0 | 0 | 0 | 1 | 0 | 0 | 0 | 0 | 0 | 0 | 0 | 6 | 131 | 2 | 0 | 0 | 0 |
| 44 | 0 | 0 | 0 | 0 | 0 | 0 | 0 | 0 | 0 | 0 | 0 | 0 | 0 | 0 | 0 | 0 | 0 | 0 | 0 | 0 | 0 | 0 | 1 | 0 | 130 | 9 | 0 | 0 |
| 45 | 0 | 0 | 0 | 0 | 0 | 0 | 0 | 0 | 0 | 0 | 0 | 0 | 0 | 0 | 0 | 0 | 0 | 1 | 0 | 0 | 0 | 1 | 0 | 0 | 10 | 127 | 1 | 0 |
| 46 | 0 | 0 | 0 | 0 | 0 | 0 | 0 | 0 | 0 | 0 | 0 | 0 | 0 | 0 | 0 | 0 | 0 | 0 | 0 | 0 | 0 | 0 | 0 | 0 | 0 | 0 | 133 | 7 |
| 47 | 0 | 0 | 0 | 0 | 0 | 0 | 0 | 0 | 0 | 0 | 0 | 1 | 1 | 0 | 0 | 0 | 0 | 0 | 0 | 0 | 1 | 0 | 0 | 0 | 0 | 1 | 5 | 131 |

## 3.2. Sex classification

Subsequent to the AI learning process, consisting of 300 male and 300 female panoramic radiographs, we then used the CNN to distinguish 120 male and 120 female panoramic radiographs not used during the learning phase. A sample of 840 patients (420 men and 420 women) was chosen randomly from 972 patients. We divided the dataset into a training set and a test set. The training set comprised 300 men and 300 women, and the test set comprised 120 men and 120 women.

### 3.2.1. Input layer

Figure 15 shows resized panoramic radiographs of 300 men and 300 women. The size of the original panoramic radiograph was 2,800 x 1,376 pixels, and the CNN training is shown as slow on large inputs. Therefore, we resized the images to 640 × 640 pixels. The 16-layer CNN model was not used due to hardware resource limitations.

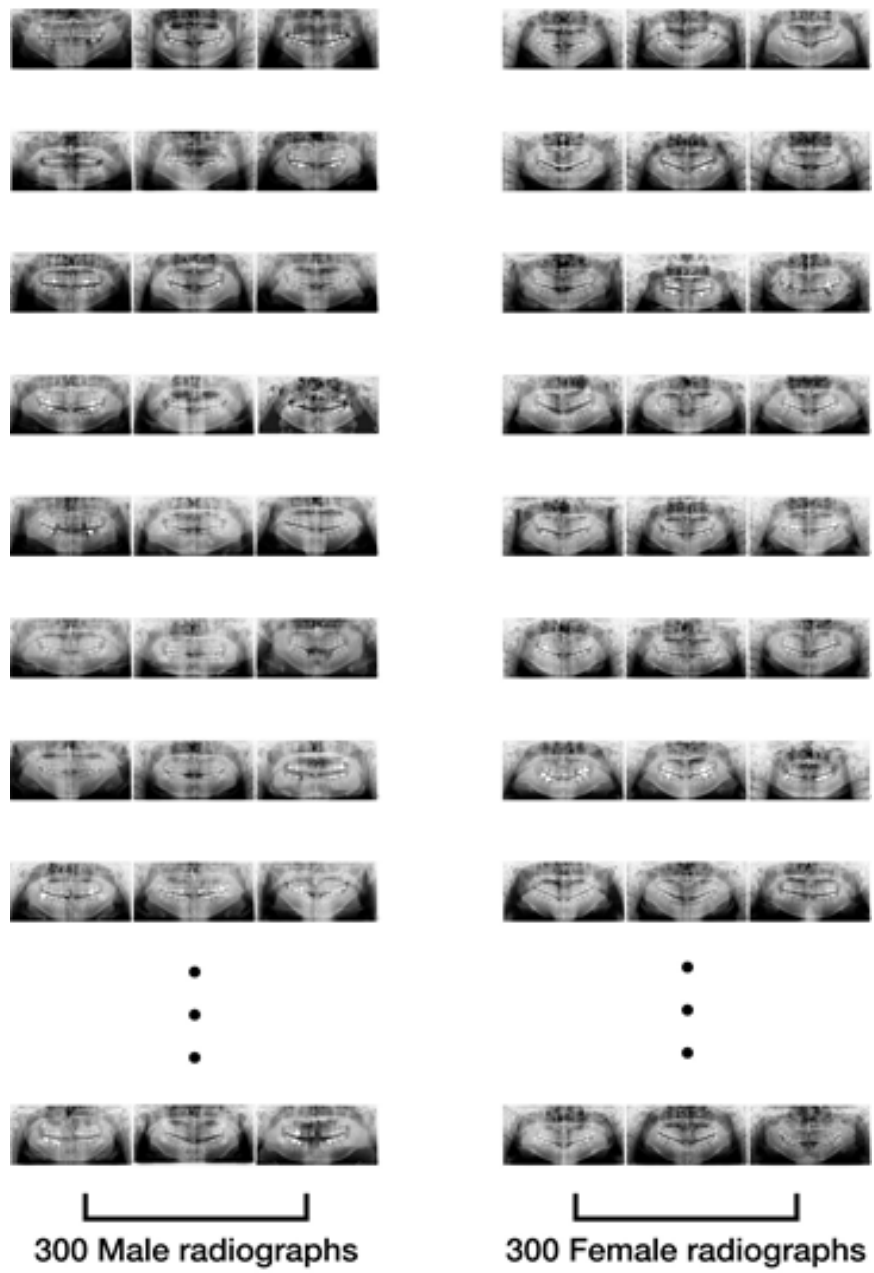

Fig. 15. Resized panoramic radiographs of 300 men and 300 women.

### 3.2.2. Tests on Bag-of-words models

As shown in Figure 16, the results indicate a maximum accuracy rate of 91.25% in the case of SIFT, 93.33% in the case of HOG, and 77.9% in the case of Color. Using these accuracy results, 7.89 s were required to distinguish 240 images with SIFT, 7.10 s with HOG, and 0.63 s with Color.

### 3.2.3. Tests on 4-layer CNN model

Figure 17 shows the changes in accuracy as a function of the number of training images using a convolutional neural network classified by a 4-layer CNN model. Training the 600 resized images required 12.70 s per number of training images, and to distinguish 240 images required 2.14 s using the 4-layer CNN model. To optimize the results of our few training examples, a number of random transformations were added to prevent each image from a second training. As shown in Table 3, training images that were processed via shear transformation showed a 1.96% higher accuracy rate than the training accuracies of the original 600 images, but there was an increase in the standard deviation. Training images processed subsequent to zoom transformation achieved the highest accuracy rate. After training, the 4-layer CNN model classified 240 test sets with a maximum accuracy rate of 97.5%.

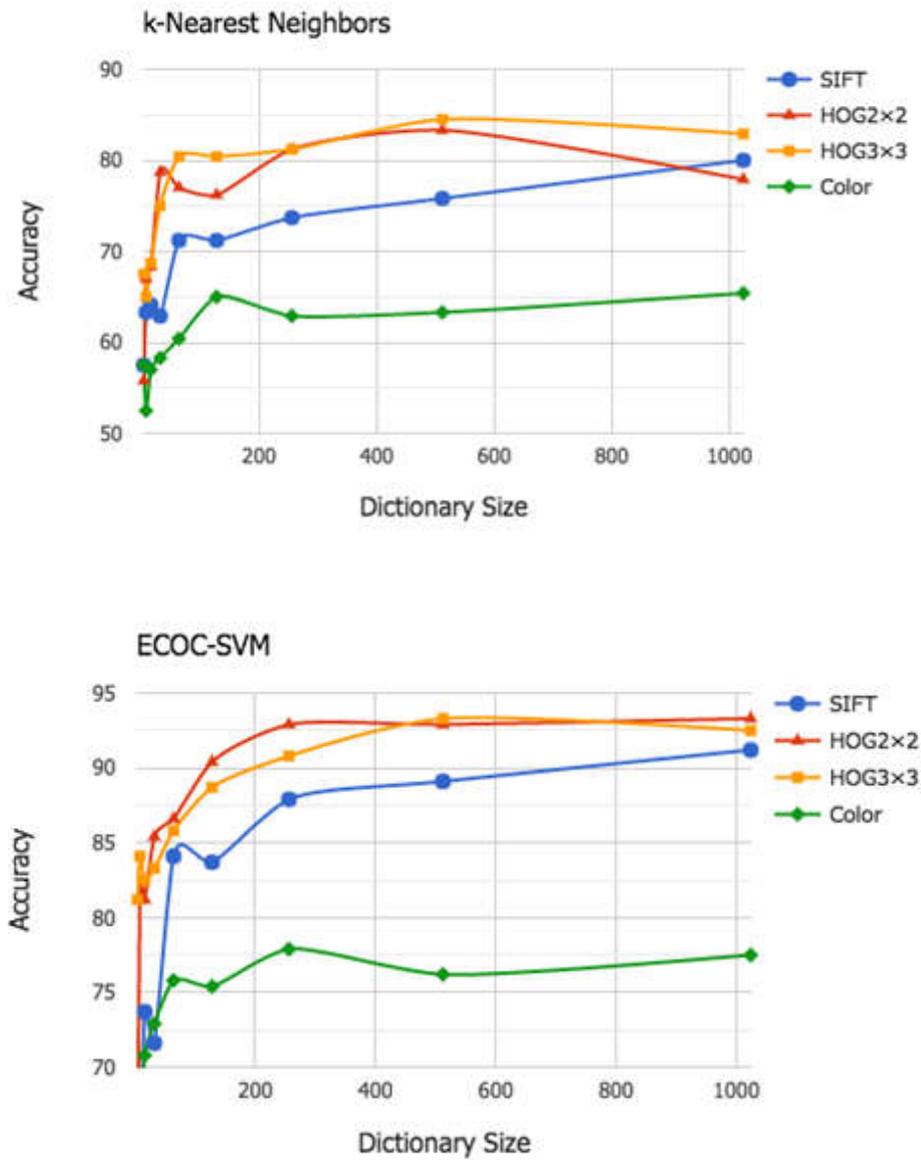

Fig. 16. Changes in accuracy with dictionary size at 640 × 640 pixels input size and level-2 pyramid using SIFT, HOG2×2, HOG3×3, and Color algorithm: (top) classified by k-NN algorithm; (bottom) classified by ECOC-SVM algorithm.

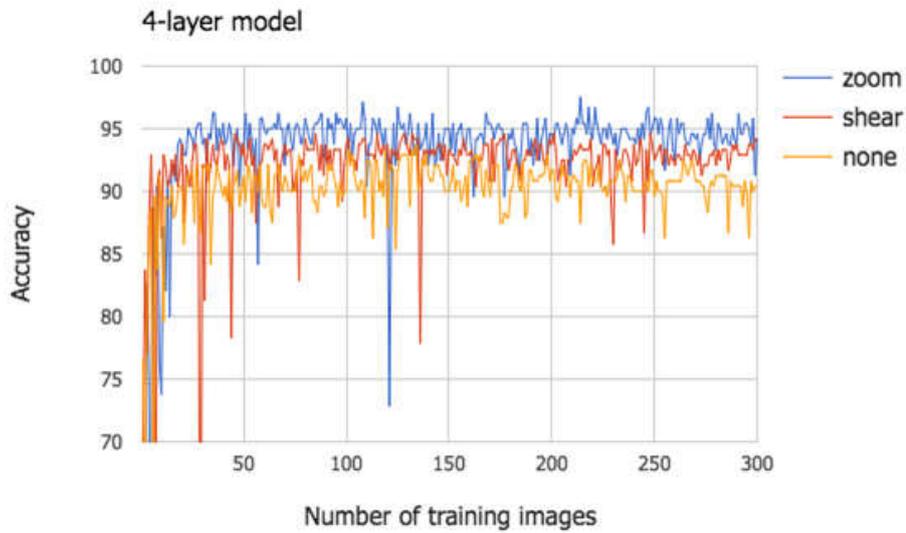

Fig. 17. Changes in accuracy with a number of training images using a convolutional neural network classified by a 4-layer CNN model: (none) original 600 resized images were used for training; (zoom) randomly zooming inside images; (shear) randomly applying shearing transformations

Table 3. Result of the performance analysis of the different methods

| Accuracy (%) | none | shear | zoom |
|---|---|---|---|
| Minimum | 86.25 | 85.83 | 91.25 |
| Maximum | 92.5 | 94.58 | 97.5 |
| Average | 90.62 | 92.66 | 94.62 |
| Standard deviation | ±1.16 | ±1.33 | ±1.07 |

## 3.3. Visualization

### 3.3.1. Visualization of HOG based object detection features

Figure 18 shows the visualization of HOG based object detection features. Since the dimensions of most feature spaces are too large to allow direct human inspection, we executed "HOGgles" algorithms to invert feature descriptors to restore these images to a natural image and to visualize feature spaces [Vondrick et al., 2013]. HOG inversions provide an accurate and intuitive visualization of feature descriptors commonly used in object detection.

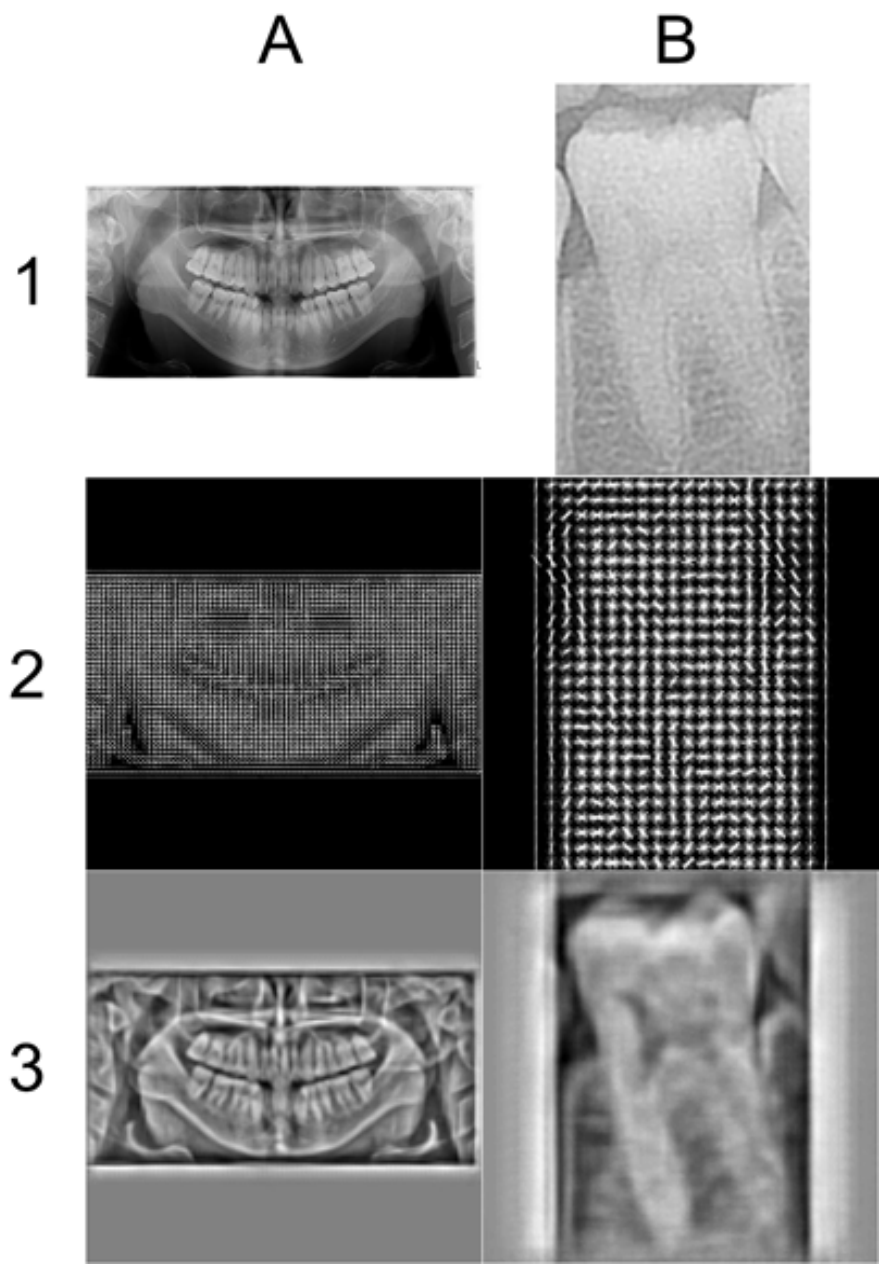

Fig. 18. Visualization of HOG based object detection features: panoramic radiography (A), molar-tooth radiography (B), original images (1), HOG features (2), HOG inversions (3).

### 3.3.2. Visualization of filters in the 4-layer CNN model

Figures 19 through 24 show the visualization of the first and second filter's weights and their activations in the teeth 4-layer classification CNN model. Several approaches for understanding and visualizing CNNs have been developed in literature, partly as a response to the common criticism that the features learned by an ANN cannot be interpreted. For example, Matthew Zeiler developed an outstanding visualization method for the CNN [Zeiler and Fergus, 2014].

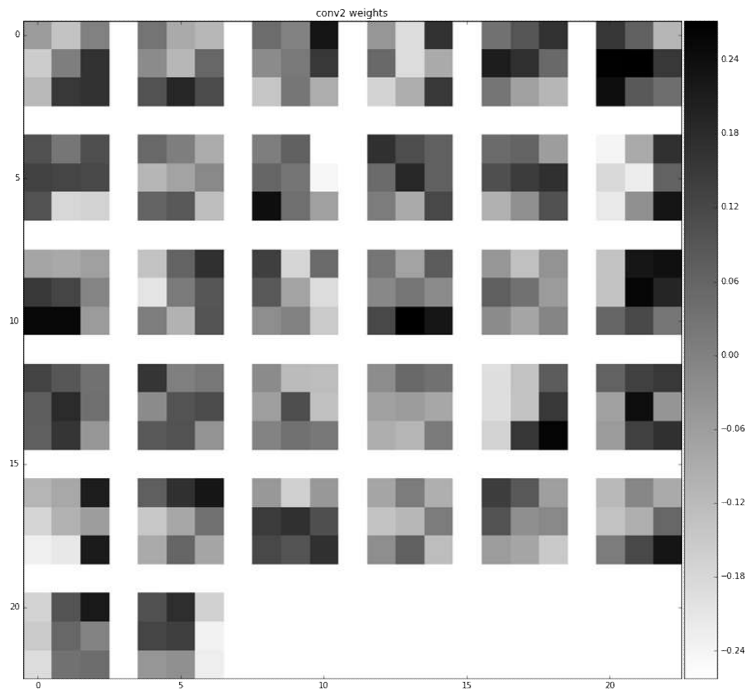

Fig. 19. Visualization of the first filter's weights after training.

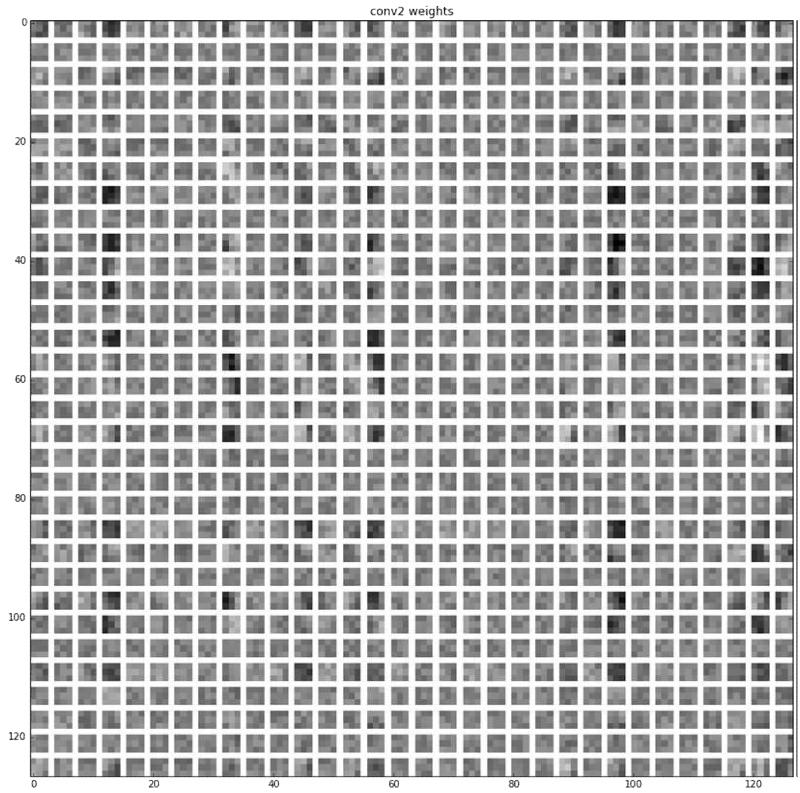

Fig. 20. Visualization of the second filter's weights after training.

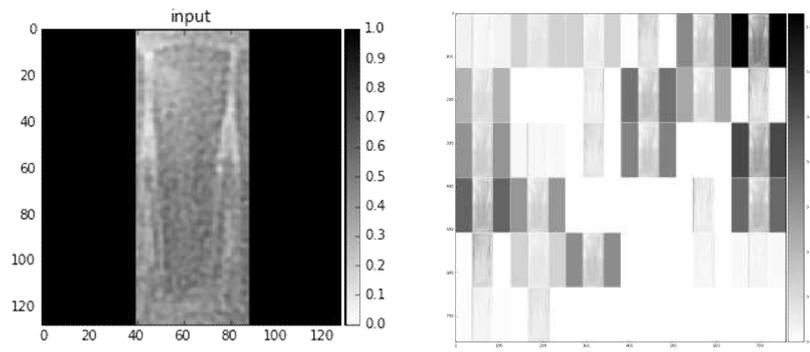

Fig. 21. Visualization of the first filter's activations with tooth #41.

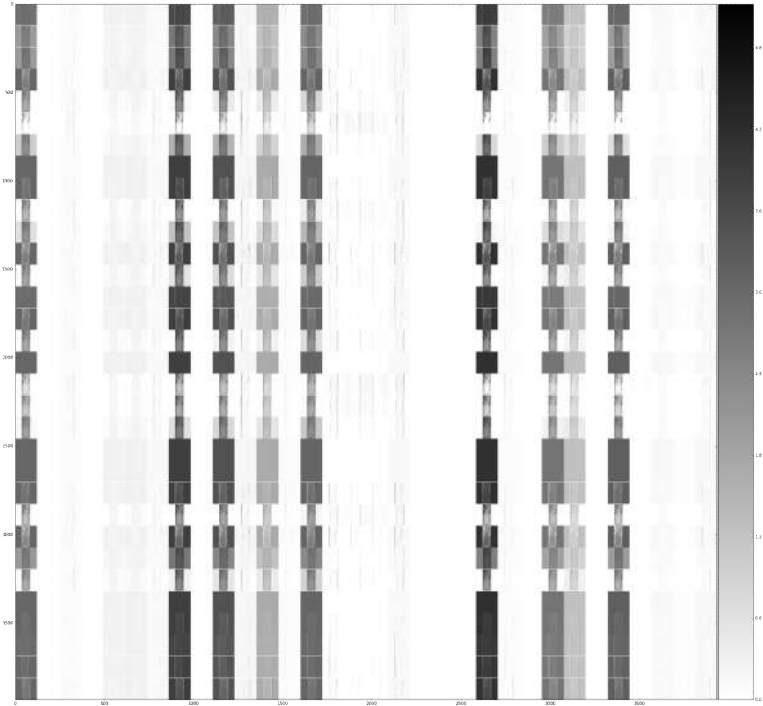

Fig. 22. Visualization of the second filter's activations with tooth #41.

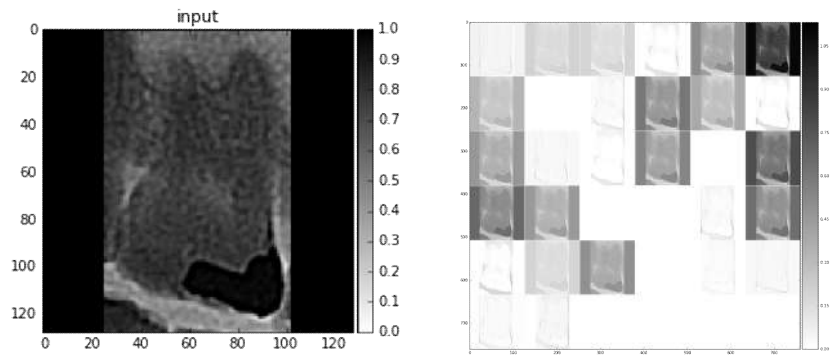

Fig. 23. Visualization of the first filter's activations with tooth #16.

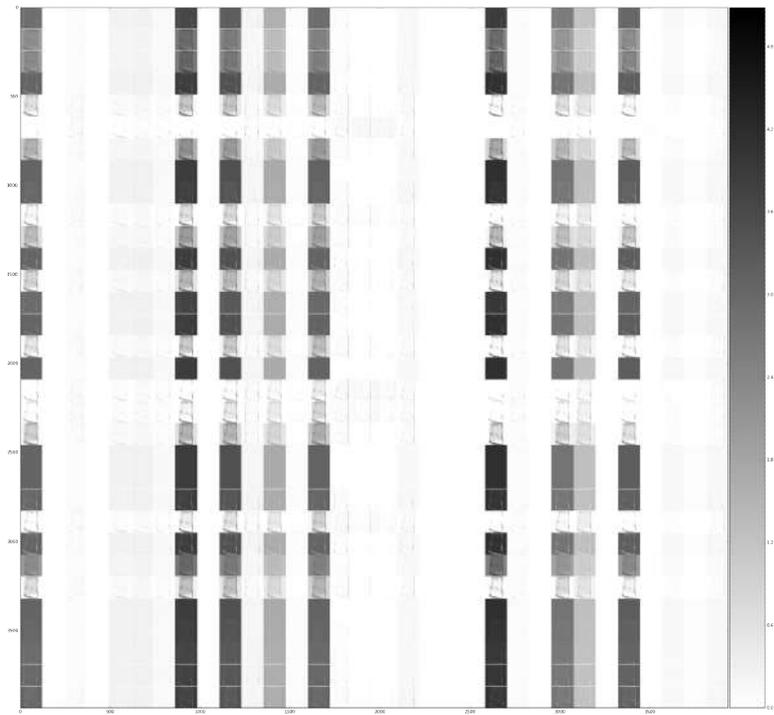

Fig. 24. Visualization of the second filter's activations with tooth #16.

# 4. Discussion

Advanced digital computer technology has dramatically assisted the practical development of panoramic radiographs, periapical radiographs, and dental computed tomography (CT). Reviewing the status of the diagnostic radiation generator in Korea in 2012, there were 15,727 diagnostic X-ray generators, 20,029 diagnostic X-ray systems, 2,618 mammography systems, 32,013 dental diagnostic X-ray systems, and 2,242 CT scanners [Kim et al., 2013]. Among these, 32,013 of the 72,626 dental diagnostic X-ray systems were used, a number that exceeds the number of mammography systems by a factor of 12. Computer-aided detection (CAD) was employed by many researchers in the detection and the classification of clustered microcalcifications on a mammogram. This was due to the very small size of the calcified lesions to detect benign and malignant breast lesions that occasionally overlap the breast tissue, making visual identification by humans difficult. Recent artificial intelligence based on artificial neural networks requires large amounts of data for better performance. The increase in the number of dental diagnostic X-ray systems and dental radiography analysis, using artificial neural networks, has created many opportunities to discover new results. One of the most common dental diagnostic X-ray systems is panoramic radiography. This technology is a very common method used by clinicians to quickly record all upper and lower teeth within a small period of time and to quickly view the entire tooth. It is easy to observe the status of the teeth regarding lesions that have and have not been cured. However, the sharpness and precision of panoramic radiographs are less relative to the periapical radiographs and dental CT. When the cervical vertebrae are overlapped by the patient's incorrect posture, the sharpness of the anterior teeth is lowered. Molander reported that mean scores for subjective image quality of the periapical bone area by the type of teeth were $2.52 \pm 0.29$ from molars, $2.49 \pm 0.23$ from premolars, and $2.37 \pm 0.30$ from (canines + incisors) using Maxilla, and $3.20 \pm 0.21$ from molars, $3.14 \pm 0.28$ from premolars, and $2.16 \pm 0.27$ from (canines + incisors) using Mandible [Molander et al., 1995]. Choi reported that averages of the regional image quality grade evaluation scores were $3.17 \pm 0.47$ from molars, $2.75 \pm 0.52$ from premolars, and $2.30 \pm 0.70$ from incisors [Choi, 2012].

In teeth classification, the maximum accuracy rate achieved was 90.36% despite the small training size. It was notable that CNN achieved this accuracy without information regarding the position of the teeth. Table 1 shows the accuracy rate of molars at 95%, premolars at 89%, and incisors at 80%. These results show the same tendency as the image quality from previous studies. The difference between the molars and the premolars, as shown in the results of the composite neural network (CNN), was greater than reported in the previous studies. It is expected that the complex structure of the molars with two or more roots influenced the learning of the convolutional neural network. A dental CT, which has less image distortion and higher image resolution than panoramic radiographs, provides more accurate results. However, dental CTs define different angles for each image, making it difficult to apply them universally to all dental CTs. The identification of teeth alone is not clinically significant. However, if the normal teeth are correctly recognized, the differences between the normal teeth and the abnormal teeth can be used to identify diseases that are difficult to find. It is also possible to categorize all the diseases, including new ones, by computing the difference in vector distance from the normal teeth. Diseases, such as Cherubism, Paget's disease of the bone, and Osteopetrosis show signs throughout the bone, including the maxilla and mandible, as well as teeth. Although these diseases can be identified by learning entire radiographs on convolutional neural networks, it is very inefficient to learn all radiographs because many dental diseases are confined to several teeth and their surroundings. It is impossible to manually determine the ROI (region of interest) in order to learn massive amounts of panoramic radiographs. Therefore, it is necessary to find the ROI automatically using the image segmentation method. Figure 25 shows the segmentation of a panoramic radiograph into regions using a graph-based representation of the image. It can be considered a way to specify the position information of teeth to improve the accuracy. Use of graph-based image segmentation or normalized cuts and image segmentation may be considered to separate teeth [Shi and Malik, 2000; Felzenszwalb and Huttenlocher, 2004]. Another limitation of this study was that the third molars were excluded in order to train marginally more samples because there are many people without third molars. Anatomical position and shape are highly variable in the third molar. For example, agenesis of the third molar differs by population, ranging from practically zero in Tasmanian Aborigines to nearly 100% in indigenous Mexicans [Nanda, 1954; Rozkovcová et al., 1998]. According to literature, the difference is related to the PAX9 gene [Pereira et al., 2006]. It would be worthwhile to study the third molars using a machine learning technique.

In sex classification, the maximum accuracy rate achieved was 97.50% using the 4-layer CNN model. The 16-layer CNN model was not used because of limitations of computing resources. Classification of gender using skulls has been studied in the forensic area. Based on the results of examining the sex features of 100 adult craniums (50 men and 50 women) through 23 items, Keen reported that sex classification by cranium was 85% accurate[Keen, 1950]. Krogman reported that the accuracy was 95% if there were only pelvic data, 90% if there were only cranial data, and 80% if there were only limb skeleton data. However, if there were cranial and pelvic data, the accuracy was 98% [Krogman, 1962]. There were also gender classification studies using radiographs of various bones. Patil and Mody reported determination of sex by discriminant function analysis and stature by regression analysis: "A Lateral Cephalometric Study." The sample space for this study, a total of 150 normal healthy adults from Central India, consisted of 75 males and 75 females. Ten cephalometric measurements (G-Op, Ba-ANS, N-ANS, Ba-N, F-M, FsHt, Ma-SN, Ma-FM, MaHt, MaWd) were used in discriminant functional analysis and they provided accurate sex discrimination in Central Indian subjects of known sex. It was observed that Ba–N, MaHt, N–M, MaWd, Ba–ANS, Ma–FH, and G–Op were major variables in the determination of sex and their respective discriminative powers were 25.88, 15.12, 13.31, 11.88, 7.78, 7.02, and 6.90% [Patil and Mody, 2005]. Indira et al. reported Mandibular ramus: "An Indicator for Sex Determination–A Digital Radiographic Study". A retrospective study was conducted using orthopantomographs of 50 males and 50 females. The overall prediction rate using all five variables (Max. ramus breadth, Min.

ramus breadth, Condylar height, Projective height of ramus, and Coronoid height) was 76% and the accuracy can be increased by repeated iterations [Indira et al., 2012]. Verma et al. reported radiomorphometric analysis of frontal sinus for sex determination. The sample space consisted of 100 patients, 50 males, and 50 females. The percentage agreement of the total area to correctly predict the female gender was 55.2%, of which, the right area was 60.9% and the left area was 55.2%, respectively [Verma et al., 2014]. The subjects of all of the studies listed above were for people over 20 years of age. It should be noted that the size of the sample is small and the methods used to classify the gender are often tested with objects that are not new objects, so errors will occur. Generally, definite sex features are shown in the skeleton from adolescence and are not revealed until after the completion of the secondary sex characteristics. The skeletal boundary between juvenile and adult is at age 15–18; among those who are younger, it is difficult to estimate the sex. It would be worthwhile to study pediatric radiography using a machine learning technique.

Section 3.3. shows visualization results of machine learning. Several approaches for understanding and visualizing CNNs have been described in the literature, partly in response to criticism regarding the fact features of an ANN have previously been reported as being difficult to understand and interpret. [Zeiler and Fergus, 2014]. We expect to find a theoretical basis through subsequent studies of inverse analysis such as a visualization method for artificial neural networks. Object recognition is a fundamental problem in computer vision that has been studied for more than 30 years. It remains a particularly difficult problem and is far from being solved. Recently, AI applications have achieved optimal performance in a variety of pattern-recognition tasks, most notably visual classification problems. However, there is a large difference between computer and human vision. The literature indicates that it is easy to produce images that are completely unrecognizable to humans, while the CNNs believe objects to be recognizable with 99.99% confidence (e.g., labeling with certainty that static white noise is a cat) [Nguyen et al., 2015]. As well as being aware of the difference between computer vision and human vision, researchers must continue to identify diseases that are easily bypassed. Deep learning has dramatically shortened the time for artificial intelligence to identify data and this also shortens the learning time. While learning more data within the domain of dentistry, we hope that artificial intelligence will be useful in this field. In addition, we hope to be able to provide useful information on chronic diseases such as osteoporosis and TMJ disorder.

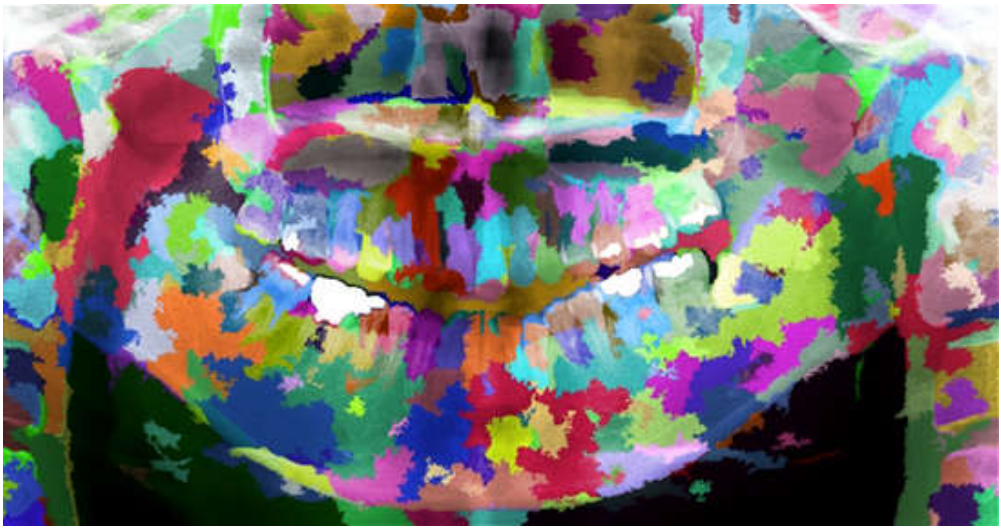

Fig. 25. Segmenting a panoramic radiograph into regions using a graph-based representation of the image.

# References

Aggarwal, M., Madhukar, M. (2016). IBM's Watson Analytics for Health Care: A Miracle Made True. *Cloud Computing Systems and Applications in Healthcare* (pp. 117-134). Hershey, PA: IGI Global.

Alipanahi, B., Delong, A., Weirauch, M. T., Frey, B. J. (2015). Predicting the sequence specificities of DNA-and RNA-binding proteins by deep learning. *Nature biotechnology*.

Behnke, S. (2003). *Hierarchical neural networks for image interpretation* (Vol. 2766). New York: Springer Science & Business Media.

Bergstra, J., Breuleux, O., Bastien, F., Lamblin, P., Pascanu, R., Desjardins, G., Turian. J., Warde-Farley. D., Bengio, Y. (2010, June). Theano: A CPU and GPU math compiler in Python. In *Proc. 9th Python in Science Conf* (pp. 1-7). Austin, TX: SciPy.

Carpenter, G. A., Grossberg, S. (2011). *Adaptive resonance theory* (pp. 22-35). New York: Springer US.

Chang, H., Choi, M. (2016). Big Data and Healthcare: Building an Augmented World. *Healthcare Informatics Research*, 22(3), 153-155.

Collobert, R., Weston, J. (2008, July). A unified architecture for natural

language processing: Deep neural networks with multitask learning. In *Proceedings of the 25th international conference on Machine learning* (pp. 160-167). ACM.

Dalal, N., Triggs, B. (2005, June). Histograms of oriented gradients for human detection. In *2005 IEEE Computer Society Conference on Computer Vision and Pattern Recognition (CVPR'05)* (Vol. 1, pp. 886-893). San Diego, CA: Institute of Electrical and Electronics Engineers.

Elkan, C. (2003, August). Using the triangle inequality to accelerate k-means. In *ICML 2003* (Vol. 3, pp. 147-153). Washington, D.C.: International Machine Learning Society.

Felzenszwalb, P. F., Huttenlocher, D. P. (2004). Efficient graph-based image segmentation. *International Journal of Computer Vision*, 59(2), 167-181.

Grossberg, S. (1976). Adaptive pattern classification and universal recoding: I. Parallel development and coding of neural feature detectors. *Biological cybernetics*, 23(3), 121-134.

Hebb, D. O. (1949). *The organization of behavior: A neuropsychological approach.* New York: John Wiley & Sons.

Herath, D. H., Wilson-Ing, D., Ramos, E., Morstyn, G. (2016, May).

Assessing the natural language processing capabilities of IBM Watson for oncology using real Australian lung cancer cases. In *ASCO Annual Meeting Proceedings* (Vol. 34, No. 15_suppl, p. e18229). Chicago, IL: American Society of Clinical Oncology.

Hinton, G., Deng, L., Yu, D., Dahl, G. E., Mohamed, A. R., Jaitly, N., Kingsbury, B. (2012). Deep neural networks for acoustic modeling in speech recognition: The shared views of four research groups. *IEEE Signal Processing Magazine*, 29(6), 82-97.

Hopfield, J. J. (1982). Neural networks and physical systems with emergent collective computational abilities. *Proceedings of the national academy of sciences*, 79(8), 2554-2558.

Indira, A. P., Markande, A., David, M. P. (2012). Mandibular ramus: An indicator for sex determination-A digital radiographic study. *Journal of forensic dental sciences*, 4(2), 58.

Johnson, B. D., Guston, D. (2016). Futures We Want to Inhabit. *Computer*, 49(2), 78-79.

Keen, J. A. (1950). A study of the differences between male and female skulls. *American Journal of Physical Anthropology*, 8(1), 65-80.

Kelemen, J. (2007). From artificial neural networks to emotion machines with marvin minsky. *Acta Polytechnica Hungarica*, 4(4), 1-12.

Khan, R., Weijer, J., Shahbaz Khan, F., Muselet, D., Ducottet, C., Barat, C. (2013). Discriminative color descriptors. In *Proceedings of the IEEE Conference on Computer Vision and Pattern Recognition* (pp. 2866-2873). Portland, OR: Institute of Electrical and Electronics Engineers.

Krogman, W. M. (1962). In sex differences in the skull, chapter 5; sexing skeletal remains. *The human skeleton in forensic medicine* (pp. 114-121). Springfield, MA: Charles C Thomas.

Lazebnik, S., Schmid, C., Ponce, J. (2006). Beyond bags of features: Spatial pyramid matching for recognizing natural scene categories. In *2006 IEEE Computer Society Conference on Computer Vision and Pattern Recognition* (CVPR'06) (Vol. 2, pp. 2169-2178). New York: Institute of Electrical and Electronics Engineers.

LeCun, Y., Bengio, Y. (1995). Convolutional networks for images, speech, and time series. *The handbook of brain theory and neural networks*, 3361(10), 1995. Cambridge, MA: MIT Press.

LeCun, Y., Boser, B., Denker, J. S., Henderson, D., Howard, R. E., Hubbard, W., Jackel, L. D. (1989). Backpropagation applied to handwritten zip code recognition. *Neural computation*, 1(4), 541-551.

Leidos, C. W., Pentagon, L. (2015). The Roundup. *Biomedical Instrumentation & Technology*, 49(5), 296-299.

Lowe, D. G. (2004). Distinctive image features from scale-invariant keypoints. *International journal of computer vision*, 60(2), 91-110.

Luszczek, P. (2009). Parallel programming in MATLAB. *International Journal of High Performance Computing Applications*, 23(3), 277-283.

McCulloch, W. S., Pitts, W. (1943). A logical calculus of the ideas immanent in nervous activity. *The bulletin of mathematical biophysics*, 5(4), 115-133.

Molander, B., Ahlqwist, M., Gröndahl, H. G. (1995). Image quality in panoramic radiography. *Dentomaxillofacial Radiology*, 24(1), 17-22.

Nair, V., Hinton, G. E. (2010). Rectified linear units improve restricted boltzmann machines. In *Proceedings of the 27th International Conference on Machine Learning (ICML-10)* (pp. 807-814). Haifa: International Machine Learning Society.

Nanda, R. S. (1954). Agenesis of the third molar in man. *American Journal of Orthodontics*, 40(9), 698-706.

Nash, D. B. (2016). Eye of the Consolidation Storm. *American health & drug benefits*, 9(6), 302-303.

Nelson, R. (2016). From Hephaestus' automatons to OpenAI's deep learning. *EE-Evaluation Engineering*, 55(2), 2-3.

Nguyen, A., Yosinski, J., Clune, J. (2015, June). Deep neural networks are easily fooled: High confidence predictions for unrecognizable images. In *2015 IEEE Conference on Computer Vision and Pattern Recognition (CVPR)* (pp. 427-436). Boston, MA: Institute of Electrical and Electronics Engineers.

Patil, K. R., Mody, R. N. (2005). Determination of sex by discriminant function analysis and stature by regression analysis: a lateral cephalometric study. *Forensic science international*, 147(2), 175-180.

Peck, S., Peck, L. (1996). Tooth numbering progress. *The Angle Orthodontist*, 66(2), 83-84.

Pereira, T. V., Salzano, F. M., Mostowska, A., Trzeciak, W. H., Ruiz-Linares, A., Chies, J. A., Saavedra, C., Nagamachi, C., Hurtado, A. M., Hill, K., Castro-de-Guerra, D., Silva-Junior, W. A., Bortolini, M. C. (2006). Natural selection and molecular evolution in primate PAX9 gene, a major determinant of tooth development. *Proceedings of the National Academy of Sciences*, 103(15), 5676-5681.

Piros, E., Petak, I., Erdos, A., John Hautman, J. D., Julianna Lisziewicz, P. (2016). Market Opportunity for Molecular Diagnostics in Personalized Cancer Therapy. *Handbook of Clinical Nanomedicine: Law, Business, Regulation, Safety, and Risk* (Vol. 2). Boca Raton, FL: CRC Press.

Ravdin, P. M., Siminoff, L. A., Davis, G. J., Mercer, M. B., Hewlett, J., Gerson, N., Parker, H. L. (2001). Computer program to assist in making decisions about adjuvant therapy for women with early breast cancer. *Journal of Clinical Oncology*, 19(4), 980-991.

Raza, M., Le, M. H., Aslam, N., Le, C. H., Le, N. T., Le, T. L. (2016). Telehealth Technology: Potentials, Challenges and Research Directions for Developing Countries. In *IFMBE Proceedings* (Vol. 20, pp. 233-236). Ho Chi Minh City: International Federation of Medical and Biological Engineering.

Rosenblatt, F. (1957). The perceptron, a perceiving and recognizing automaton *Project Para*(Report No. 85-460-1). New York: Cornell Aeronautical Laboratory.

Rozkovcová, E., Markova, M., Dolejsi, J. (1998). Studies on agenesis of third molars amongst populations of different origin. *Sbornik lekarsky*, 100(2), 71-84.

Russell, B. C., Torralba, A., Murphy, K. P., Freeman, W. T. (2008). LabelMe: a database and web-based tool for image annotation. *International journal of computer vision*, 77(1-3), 157-173.

Shi, J., Malik, J. (2000). Normalized cuts and image segmentation. *IEEE Transactions on pattern analysis and machine intelligence*, 22(8), 888-905.

Silver, D., Huang, A., Maddison, C. J., Guez, A., Sifre, L., Van Den Driessche, G., Schrittwieser, J., Antonoglou, I., Panneershelvam, V., Lanctot, M., Dieleman, S., Nham, J., Kalchbrenner, N., Sutskever, I., Lillicrap, T., Leach, M., Kavukcuoglu, K., Graepel, T., Hassabis, D. (2016). Mastering the game of Go with deep neural networks and tree search. *Nature*, 529(7587), 484-489.

Simard, P. Y., Steinkraus, D., Platt, J. C. (2003, August). Best practices for convolutional neural networks applied to visual document analysis. In *ICDAR 2003* (Vol. 3, pp. 958-962). Edinburgh: International Conference on Document Analysis and Recognition.

Simonyan, K., Zisserman, A. (2014). Very deep convolutional networks for large-scale image recognition. *arXiv preprint* arXiv:1409.1556.

Tan, J. W., Andrade, A. O., Li, H., Walter, S., Hrabal, D., Rukavina, S., Limbrecht-Ecklundt, K., Hoffman, H., Traue, H. C. (2016). Recognition of Intensive Valence and Arousal Affective States via Facial Electromyographic Activity in Young and Senior Adults. PloS one, 11(1), e0146691.

Weijer, J., Schmid, C., Verbeek, J., Larlus, D. (2009). Learning color names for real-world applications. *IEEE Transactions on Image Processing*, 18(7), 1512-1523.

Verma, S., Mahima, V. G., Patil, K. (2014). Radiomorphometric analysis of frontal sinus for sex determination. *Journal of forensic dental sciences*, 6(3), 177.

Vondrick, C., Khosla, A., Malisiewicz, T., Torralba, A. (2013). Hoggles: Visualizing object detection features. In Proceedings of the IEEE International Conference on Computer Vision (pp. 1-8). Sidney: Institute of Electrical and Electronics Engineers.

Wang, J., Yang, J., Yu, K., Lv, F., Huang, T., Gong, Y. (2010, June). Locality-constrained linear coding for image classification. In *Computer Vision and Pattern Recognition (CVPR)*, 2010 IEEE Conference on (pp. 3360-3367). San Francisco, CA: Institute of Electrical and Electronics Engineers.

Williams, D. R. G. H. R., Hinton, G. E. (1986). Learning representations by back-propagating errors. *Nature*, 323, 533-536.

Winsberg, F., Elkin, M., Macy Jr, J., Bordaz, V., Weymouth, W. (1967). Detection of radiographic abnormalities in mammograms by means of optical scanning and computer analysis 1. *Radiology*, 89(2), 211-215.

Zeiler, M. D., Fergus, R. (2014, September). Visualizing and understanding convolutional networks. In *European Conference on Computer Vision* (pp. 818-833). Zürich: Springer International Publishing.